\documentclass{article}

\usepackage[preprint]{staix_2026}

\usepackage[utf8]{inputenc}
\usepackage[T1]{fontenc}
\usepackage{microtype}
\usepackage{amsmath, amssymb, amsthm, mathtools}
\usepackage{bm}
\usepackage{booktabs}
\usepackage{graphicx}
\usepackage{subcaption}
\usepackage{enumitem}
\usepackage[table]{xcolor}
\usepackage{hyperref}
\usepackage[capitalize]{cleveref}
\usepackage{tikz}
\usetikzlibrary{arrows.meta, positioning, fit, calc, shapes.geometric, decorations}
\usepackage{algorithm}
\usepackage{url}

\definecolor{obsfill}{HTML}{F1EFE8}
\definecolor{obsstroke}{HTML}{888780}
\definecolor{auditfill}{HTML}{B5D4F4}
\definecolor{auditstroke}{HTML}{185FA5}
\definecolor{goldfill}{HTML}{FAC775}
\definecolor{goldstroke}{HTML}{854F0B}
\definecolor{mfill}{HTML}{CECBF6}
\definecolor{mstroke}{HTML}{534AB7}
\definecolor{yfill}{HTML}{F7C1C1}
\definecolor{ystroke}{HTML}{A32D2D}
\definecolor{bandfill}{HTML}{85B7EB}
\definecolor{textprimary}{HTML}{2C2C2A}
\definecolor{textsecondary}{HTML}{5F5E5A}

\usetikzlibrary{decorations.pathmorphing} 
\usetikzlibrary{decorations.pathreplacing} 
\usetikzlibrary{decorations.text}          
\usetikzlibrary{decorations.markings}      

\hypersetup{
  colorlinks=true,
  linkcolor=blue!50!black,
  citecolor=blue!50!black,
  urlcolor=blue!50!black
}

\newcommand{\R}{\mathbb{R}}
\newcommand{\E}{\mathbb{E}}
\newcommand{\PR}{\mathbb{P}}
\newcommand{\one}{\mathbf{1}}
\newcommand{\X}{\bm{X}}
\newcommand{\bbeta}{\bm{\beta}}
\newcommand{\Ystar}{Y^{\star}}
\newcommand{\expit}{\operatorname{expit}}
\newcommand{\PADSL}{\textsc{PA-DSL}}

\newcommand{\indep}{\perp\!\!\!\perp}

\newcommand{\best}[1]{\bm{#1}}                                    
\newcommand{\oraclesep}{\arrayrulecolor{black!40}\midrule\arrayrulecolor{black}}

\newtheorem{proposition}{Proposition}
\newtheorem{assumption}{Assumption}

\crefname{assumption}{Assumption}{Assumptions}
\Crefname{assumption}{Assumption}{Assumptions}
\crefname{proposition}{Proposition}{Propositions}
\Crefname{proposition}{Proposition}{Propositions}

\title{Design-Based Supervised Learning \\ with Noisy Human Labels}

\author{%
  Robert Chew \\
  RTI International
  \and
  \textbf{Matthew R. Williams} \\
  U.S. Bureau of Labor Statistics
}

\begin{document}

\maketitle

\begin{abstract}
Researchers increasingly use automated classifiers to label unstructured data for statistical analysis. Existing rectification methods can correct errors in these automated labels using a probability-sampled audit set, but they usually treat the audit labels as correct. In practice, human audit labels are often noisy, and only some audited items are reviewed by an expert or adjudicator. We propose Partially Adjudicated Design-Based Supervised Learning (\PADSL{}), a method for this setting. It uses adjudicated cases to correct noisy human labels and then uses the corrected audit information to debias analyses based on the full set of automated labels. The estimator is valid for a broad class of downstream analyses when the audit and adjudication probabilities are known. In synthetic and Wikipedia Detox semi-synthetic experiments, \PADSL{} maintains nominal coverage and reduces RMSE by 10--17\% relative to using only adjudicated labels when noisy human labels contain recoverable signal.
\end{abstract}

\section{Introduction}
\label{sec:intro}

Empirical research in the social sciences, business, law, and related fields often depends on coding raw texts, images, or records into variables that measure substantive constructs \citep{krippendorff2018content, grimmer2013text}. Traditionally, this coding has relied on human or expert judgment, making it costly, time-consuming, and prone to inconsistency. Automated classifiers now make it possible to code data at scale, ranging from traditional supervised text classifiers to newer LLM-based systems \citep{gilardi2023chatgpt, chew2023llm, tornberg2025large}. But classifier-generated labels are noisy measurements, not observed outcomes \citep{wang2020methods}. Systematic prediction errors can bias downstream estimates and produce undercovered confidence intervals, even when classifiers achieve high overall accuracy \citep{neuhaus1999bias}.

This has motivated a growing set of rectification methods that use higher-quality validation or audit data to correct estimates based on full-population surrogate labels \citep{hopkins2010method, wang2020methods, angelopoulos2023prediction, egami2023using, fong2021machine}. The common structure is simple: inexpensive automated labels provide scale, while a smaller set of trusted labels anchors the correction. Design-Based Supervised Learning (DSL), for example, combines a full-population surrogate score with a probability-sampled audit to obtain valid inference for downstream estimands such as regression coefficients \citep{egami2023using}.

In many real labeling pipelines, however, the audit labels used for correction are not themselves gold standard. Human annotators disagree, and disagreement often reflects ambiguity, subjectivity, or guideline interpretation rather than only random coder error \citep{aroyo2015truth,uma2021learning, chew2026ground}.
Because expert reconciliation is costly \citep{snow2008cheap,chau2020cost}, a common workflow is to have one or more human annotators label all items from an ``audited'' sample, with only selected items from it receiving expert adjudication. This multi-reviewer or adjudication design is widespread, reappearing across domains diverse as content analysis, systematic reviews, clinical endpoint assessment, and NLP annotation \citep{lombard2002content,page2021prisma,held2019we,finlayson2017overview}.
Existing rectification strategies do not fully address this setting because the audit used to correct the surrogate may itself be measured with error.

This paper introduces \emph{Partially Adjudicated Design-Based Supervised Learning} (\PADSL{}), a design-based estimator for downstream inference under hierarchical label noise. Conceptually, \PADSL{} is not a new estimating principle from scratch, but a nested application of known-probability AIPW/DSL corrections to a partially adjudicated annotation design. It uses an inner correction to turn the adjudicated subset into audit-level pseudo-labels, and an outer correction to use those pseudo-labels to debias inference based on the full-population surrogate. To our knowledge, this is the first work to extend rectification methods beyond the usual gold-audit setting, allowing the human audit labels used for correction to be noisy and only partially adjudicated.

\paragraph{Contributions.}
The paper makes three contributions. First, it formalizes a three-tier nested measurement design with noisy audit labels and partial adjudication. Second, it proves that the resulting pseudo-outcome is design-valid for any downstream estimating equation affine in the latent outcome, provided the audit and adjudication probabilities are known by design. Third, it shows in synthetic and semi-synthetic experiments that \PADSL{} retains the coverage of adjudication-only DSL while improving efficiency when the noisy audit labels contain recoverable information about the adjudicated construct.

\section{Setup: Hierarchical Label Noise and Target Estimand}
\label{sec:setup}

\paragraph{Observed data.}
We observe an i.i.d.\ sample of size $N$, where each unit $i$ contributes
\begin{equation}
  W_i \;=\; \bigl(\X_i,\, Q_i,\, R_i,\, G_{i,1},\, G_{i,2},\, V_i,\, A_i\bigr),
\end{equation}
generated jointly with a latent binary truth $\Ystar_i \in \{0,1\}$ that the analyst does \emph{not} observe. Here $\X_i \in \R^{p+1}$ is a covariate vector with leading intercept; $Q_i \in [0,1]$ is a real-valued surrogate
score; $R_i \in \{0,1\}$ is the audit indicator with known sampling probability $\pi_i \in (0,1)$; $G_{i,1}, G_{i,2} \in \{0,1\}$ are two human coder labels observed iff $R_i = 1$; $V_i \in \{0,1\}$ is the adjudication indicator with known probability $\rho_i \equiv \PR(V_i = 1 \mid R_i = 1, Z_i) \in (0,1)$ a known function of the inner features defined just below; and $A_i \in \{0,1\}$ is the adjudicated label, observed iff $R_i V_i = 1$. The score \(Q_i\) need not be calibrated; it is treated as an observed surrogate feature whose systematic errors are corrected by the audit design. We
write $B_i = (Q_i, \X_i^\top)^\top$ for the \emph{outer} feature vector and $Z_i = (Q_i, \X_i^\top, G_{i,1}, G_{i,2})^\top$ for the \emph{inner} feature vector. We use two coder labels as the running case because this is a common annotation workflow and makes disagreement-based adjudication easy to describe. However, the construction does not require exactly two labels: more generally, \(Z_i\) may contain any audit-tier information observed when \(R_i=1\), including \(K_i\) coder labels, coder identities, agreement indicators, vote shares, uncertainty scores, item metadata, or other features used in the adjudication design.
The probabilities $\pi_i$ and $\rho_i$ are design parameters chosen by the researcher, so they are \emph{known}, not estimated.

\begin{figure}[t]
\centering
\begin{tikzpicture}[
  x=1mm, y=-1mm,
  every node/.style={inner sep=0pt, outer sep=0pt},
  obs/.style    ={draw=obsstroke,   fill=obsfill,   line width=0.4pt, rounded corners=0.6pt},
  auditcol/.style={draw=auditstroke, fill=auditfill, line width=0.4pt, rounded corners=0.6pt},
  gold/.style   ={draw=goldstroke,  fill=goldfill,  line width=0.4pt, rounded corners=0.6pt},
  mcol/.style   ={draw=mstroke,     fill=mfill,     line width=0.4pt, rounded corners=0.6pt},
  ycol/.style   ={draw=ystroke,     fill=yfill,     line width=0.4pt, rounded corners=0.6pt},
  miss/.style   ={draw=obsstroke,   line width=0.2pt, rounded corners=0.6pt, dashed, opacity=0.35},
  pipe/.style   ={-{Stealth[length=1.4mm, width=1.2mm]}, line width=0.5pt, draw=textsecondary},
  scorearr/.style={-{Stealth[length=1.4mm, width=1.2mm]}, line width=0.5pt, draw=textsecondary,
                   decorate, decoration={snake, amplitude=0.4mm, segment length=2mm, post length=1.2mm}},
  zoom/.style   ={draw=auditstroke, line width=0.25pt, dashed, opacity=0.55},
  steplabel/.style={font=\footnotesize\bfseries, text=textprimary},
  sublabel/.style ={font=\scriptsize, text=textsecondary},
  collabel/.style ={font=\scriptsize, text=textprimary},
  betalabel/.style={font=\small, text=textprimary},
]

\node[steplabel] at (16, 0) {Step 1};
\node[sublabel]  at (16, 4) {Three-tier Sample};

\fill[bandfill, opacity=0.18, rounded corners=0.4pt] (12, 14) rectangle (32, 20);

\draw[obs]      (0,  8)  rectangle (5,  32);
\draw[obs]      (6,  8)  rectangle (11, 32);
\node[collabel] at (2.5, 20) {$Q$};
\node[collabel] at (8.5, 20) {$X$};

\draw[auditcol] (13, 14) rectangle (18, 20);
\draw[auditcol] (19, 14) rectangle (24, 20);
\node[collabel] at (15.5, 17) {$G_1$};
\node[collabel] at (21.5, 17) {$G_2$};
\draw[miss] (13, 8)  rectangle (24, 13);
\draw[miss] (13, 21) rectangle (24, 32);

\draw[gold]     (26, 14) rectangle (31, 17);
\node[collabel] at (28.5, 15.5) {$A$};
\draw[miss] (26, 8)  rectangle (31, 13);
\draw[miss] (26, 18) rectangle (31, 32);

\draw[zoom] (32, 14) -- (39, 8);
\draw[zoom] (32, 20) -- (39, 22);

\node[steplabel] at (64, 0) {Step 2};
\node[sublabel]  at (64, 4) {Inner Pass (Eq.~3)};

\draw[obs]      (39, 8)  rectangle (44, 22);
\draw[obs]      (45, 8)  rectangle (50, 22);
\draw[auditcol] (51, 8)  rectangle (56, 22);
\draw[auditcol] (57, 8)  rectangle (62, 22);
\draw[gold]     (63, 8)  rectangle (68, 15);
\draw[miss]     (63, 16) rectangle (68, 22);
\node[collabel] at (41.5, 15) {$Q$};
\node[collabel] at (47.5, 15) {$X$};
\node[collabel] at (53.5, 15) {$G_1$};
\node[collabel] at (59.5, 15) {$G_2$};
\node[collabel] at (65.5, 12) {$A$};

\draw[pipe] (70, 15) -- (75, 15);

\draw[mcol] (76, 8) rectangle (83, 22);
\node[collabel] at (79.5, 15) {$\widehat{M}$};


\node[steplabel] at (111, 0) {Step 3};
\node[sublabel]  at (111, 4) {Outer Pass: DSL on $\widehat{M}$ (Eq.~4)};

\draw[obs]      (90, 8)  rectangle (95,  32);
\draw[obs]      (96, 8)  rectangle (101, 32);
\node[collabel] at (92.5,  20) {$Q$};
\node[collabel] at (98.5,  20) {$X$};

\draw[mcol]     (103, 14) rectangle (109, 20);
\node[collabel] at (106, 17) {$\widehat{M}$};
\draw[miss]     (103, 8)  rectangle (109, 13);
\draw[miss]     (103, 21) rectangle (109, 32);

\draw[pipe] (110, 20) -- (114, 20);

\draw[ycol]     (115, 8)  rectangle (121, 32);
\node[collabel] at (118, 20) {$\widetilde{Y}$};

\draw[scorearr] (122, 20) -- (128, 20);
\node[betalabel] at (131, 20) {$\widehat{\beta}$};

\end{tikzpicture}
\caption{PA-DSL pipeline. Step 1 shows the three nested measurement tiers. Every unit has covariates and an automated score \((X,Q)\) (Tier 1, gray); a randomized audit sample with known probability \(\pi_i\) adds two noisy human labels \((G_1,G_2)\) (Tier 2, blue); and a randomized subset of audited units with known probability \(\rho_i\) receives an adjudicated label \(A\) (Tier 3, amber). Bar heights encode measurement tier, and dashed outlines mark labels that are unobserved. Step 2 is the new inner correction, which uses the adjudicated cases to turn the noisy audit information into an audit-tier pseudo-label \(\widehat M\) (purple). Step 3 applies standard DSL to \(\widehat M\), producing the final pseudo-outcome \(\widetilde Y\) (coral) used to estimate \(\widehat\beta\).}
\label{fig:pa-dsl-pipeline}
\end{figure}
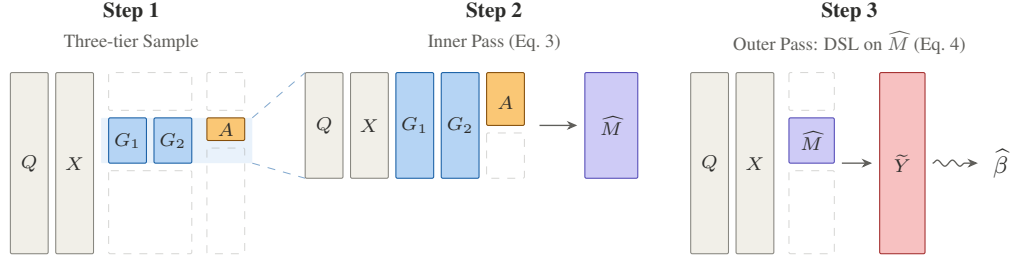

\Cref{fig:pa-dsl-pipeline} summarizes the nested three-tier sampling design. Throughout, the tiers refer to nested measurement availability, not mutually exclusive samples. Tier 1 contains all $N$ units and includes $(X_i,Q_i)$. Tier 2 is the audit subset $\{i:R_i=1\}$, which also includes noisy coder labels $(G_{i1},G_{i2})$. Tier 3 is the adjudicated subset $\{i:R_iV_i=1\}$, which also includes the adjudicated label $A_i$. Hence tier 3 $\subset$ tier 2 $\subset$ tier 1.

\paragraph{Target estimand.}
In the applications motivating this paper, the scientific target is often not whether each individual text or record is classified correctly, but how a target construct varies across groups, treatments, or covariates. For subjective annotation tasks, we interpret \(\Ystar_i\) as the label that would be assigned to unit \(i\) under the target adjudication protocol, rather than as an externally verified objective truth. We therefore target a population-level regression summary for this adjudication-defined construct: the logistic projection of \(\Ystar_i\) onto the covariates. Formally, we focus on the parameter $\beta \in \R^{p+1}$ defined as the unique solution of the population score equation
\begin{equation}
  \E\bigl[\X_i\,\bigl(\Ystar_i - \expit(\X_i^\top \beta)\bigr)\bigr]
  \;=\; \mathbf{0},
  \label{eq:beta-pop}
\end{equation}
reporting the slope $\beta_1$ on the first non-intercept covariate. This is the logistic-regression M-estimand for $\Ystar$ given $\X$. Validity for an objective construct beyond the protocol-defined target requires the additional measurement-validity assumption that the adjudication protocol itself captures that construct.

The working model serves only as a specification choice for the score equation; it has a literal conditional-probability interpretation if and only if it holds exactly, which we
do not assume. \Cref{eq:beta-pop} is \emph{outcome-affine}: the score is linear in $\Ystar_i$ for any fixed $\beta$. The construction in Section~\ref{sec:estimator} produces an unbiased pseudo-outcome $\widetilde Y_i$ for $\Ystar_i$, so any outcome-affine estimating equation (the population mean, generalized linear regression, inverse-propensity weighted treatment effects, etc.) can be solved by substituting $\widetilde Y_i$ for $\Ystar_i$. We treat the population prevalence $\E[\Ystar]$ as the special case in which the score is $\Ystar_i - \psi$; see \cref{appx:prevalence} for that variant and its empirical results.

Importantly, we do not assume that $Q_i$ is calibrated, unbiased, or a consistent estimator of $Y_i^\star$. The surrogate score is treated as an observed feature available on the full frame. Its systematic errors are corrected by the audit/adjudication design, as in standard DSL. Identification comes from the known audit and adjudication probabilities together with the ignorability/correctness assumptions below, not from any correctness property of $Q_i$. The quality of $Q_i$ affects efficiency through the nuisance functions, but not design validity.

\paragraph{Identifying assumptions.}
The audit/adjudication design supplies four assumptions; together they
identify the population means we will need.

\begin{assumption}[Audit ignorability]
  \label{ass:audit}
  $R_i \indep \Ystar_i \mid (\X_i, Q_i)$, with $\pi_i$ bounded away from
  $0$ and $1$. Holds by construction whenever audit assignment is a
  randomized function of observed design variables.
\end{assumption}

\begin{assumption}[Adjudication ignorability]
  \label{ass:adj}
  $V_i \indep \Ystar_i \mid (R_i = 1, Z_i)$, with $\rho_i$ bounded away from
  $0$ and $1$. Holds whenever adjudication is randomized with known probability as a known function of $Z_i$ --- for example, high-probability adjudication on coder disagreement and lower but nonzero adjudication on agreement cases. Strictly deterministic prioritization rules require an additional positive sampling floor in every design stratum used for inference (\cref{appx:design-curves} sweeps this floor).
\end{assumption}

\begin{assumption}[Protocol-level Adjudication correctness]
  \label{ass:correct}
On adjudicated units, \(A_i=Y_i^\star\), where \(Y_i^\star\) denotes the label that would be assigned under the target adjudication protocol.
\end{assumption}

\begin{assumption}[Coder labels are observed noisy features]
  \label{ass:coders}
The coder labels \((G_{i,1},G_{i,2})\) and any other audit-tier metadata included in \(Z_i\) are observed on the audit set and may be arbitrary noisy functions of the latent truth, item difficulty, coder identity, task context, or other factors. We do not require coder independence, unbiasedness, a fixed number of coders, or a parametric coder-error model.
\end{assumption}

Assumption 4 is a measurement condition rather than an identifying model: All audit-tier information, including \((G_{i,1},G_{i,2})\), enter only as observed features in \(Z_i\). Their bias, dependence, and heterogeneity are absorbed into \(\mu_0(Z_i)\); design validity
comes from the known adjudication probability \(\rho_i\) and adjudication
correctness.

\section{Estimator: Partially Adjudicated DSL}
\label{sec:estimator}

Standard DSL \citep{egami2023using} replaces the
unobserved $\Ystar_i$ with a pseudo-outcome of the form
$\hat g(B_i) + (R_i / \pi_i)\bigl(Y_i^{\text{gold}} - \hat g(B_i)\bigr)$, where $\hat g$ is a cross-fit nuisance regression of the gold label on $B_i$ and the Horvitz--Thompson correction debiases that regression at the known
audit weight. The construction assumes $Y_i^{\text{gold}} = \Ystar_i$ on the audit set; when the audit labels are themselves noisy, substituting any one of $G_{i,1}$, $G_{i,2}$, or their majority vote inherits the human bias. We address this by nesting a second design-based correction inside, exploiting the fact that the
adjudicated label $A_i$ is unbiased for $\Ystar_i$ on $\{R_i V_i = 1\}$ with known sampling weight $\rho_i$.

\paragraph{Step 1 --- Inner correction.}
Given any cross-fit estimator $\hat\mu(\cdot)$ of the inner nuisance $\mu_0(z) = \E[A_i \mid Z_i = z, R_i = 1, V_i = 1]$ trained on the adjudicated subset $\{R_i V_i = 1\}$, define the audit-level pseudo-label
\begin{equation}
  \widehat M_i \;=\; \hat\mu(Z_i)
    \;+\; \frac{V_i}{\rho_i}\,\bigl(A_i - \hat\mu(Z_i)\bigr),
  \qquad i \in \{R_i = 1\}.
  \label{eq:Mhat}
\end{equation}
Here $\hat\mu(Z_i)$ uses the surrogate, the covariates, and the two coder labels to predict adjudicated truth, and the $V_i/\rho_i$ term debiases that prediction on the adjudicated subset. This definition is unchanged if \(Z_i\) contains more than two coder labels or additional audit-tier metadata; the inner correction only requires that the adjudication probability \(\rho_i=P(V_i=1\mid R_i=1,Z_i)\) be known by design and bounded away from zero.

\paragraph{Step 2 --- Outer correction.}
Given a cross-fit estimator $\hat g(\cdot)$ of the outer nuisance
$g_0(b) = \E\{\mu_0(Z_i)\mid B_i=b,R_i=1\}$
trained on the audit set
$\{R_i = 1\}$ using $\widehat M_i$ as response, define the final
pseudo-outcome
\begin{equation}
  \widetilde Y_i \;=\; \hat g(B_i)
    \;+\; \frac{R_i}{\pi_i}\,\bigl(\widehat M_i - \hat g(B_i)\bigr),
  \qquad i = 1, \dots, N.
  \label{eq:Ytilde}
\end{equation}
Under Assumptions~1--3, this target satisfies \(g_0(B_i)=\E(Y_i^\star\mid B_i)\); the realized pseudo-label \(\widehat M_i\) is used only to estimate this population nuisance.

Substituting \cref{eq:Mhat} into \cref{eq:Ytilde} gives the closed form below, which makes the three roles transparent
\begin{equation}
  \boxed{\;
  \widetilde Y_i \;=\;
    \underbrace{\hat g(B_i)}_{\text{full-frame regression}}
    \;+\; \underbrace{\frac{R_i}{\pi_i}\,\bigl[\hat\mu(Z_i) - \hat g(B_i)\bigr]}_{\text{audit correction}}
    \;+\; \underbrace{\frac{R_i V_i}{\pi_i \rho_i}\,\bigl[A_i - \hat\mu(Z_i)\bigr]}_{\text{adjudication correction}}
  \;}
  \label{eq:Ytilde-expanded}
\end{equation}

Thus, PA-DSL uses adjudication only to correct the part of the audit information
that cannot be trusted, while still allowing the unadjudicated human labels to
improve efficiency through \(\hat\mu(Z_i)\).

\paragraph{Step 3 --- Downstream estimating equation.}
Substitute $\widetilde Y_i$ for $\Ystar_i$ in the population score
\cref{eq:beta-pop} and solve the empirical analog:
\begin{equation}
  \frac{1}{N} \sum_{i=1}^{N}
    \X_i\,\bigl(\widetilde Y_i - \expit(\X_i^\top \hat\beta)\bigr) \;=\; \mathbf{0},
  \label{eq:beta-emp}
\end{equation}
returning the cross-fit estimator $\hat\beta$. Standard errors come from the
sandwich formula evaluated at $(\X_i, \widetilde Y_i)$.

\paragraph{Why \texorpdfstring{$\widetilde Y_i$}{Y-tilde} is design-valid.}
Both corrections in \cref{eq:Mhat,eq:Ytilde} are AIPW-style augmentations \citep{robins1994estimation} with known design weights. Under \crefrange{ass:audit}{ass:correct}, iterated expectations give
\(\E[\widehat M_i\mid R_i=1,Z_i]=\E[\Ystar_i\mid R_i=1,Z_i]\) and \(\E[\widetilde Y_i\mid B_i]=\E[\Ystar_i\mid B_i]\), regardless of nuisance specification. Thus the nuisances affect efficiency, not design validity; full
algebra and influence functions are in \cref{appx:theory}.

\begin{algorithm}[tb]
  \caption{\PADSL{} cross-fit pseudo-outcome.}
  \label{alg:ngdsl}
  \textbf{Input:} data $\{(\X_i, Q_i, R_i, G_{i,1}, G_{i,2}, V_i, A_i)\}_{i=1}^N$;
  design weights $\{\pi_i, \rho_i\}_{i=1}^N$; folds $K$; nuisance learners
  for $\hat\mu, \hat g$.\\
  \textbf{Output:} pseudo-outcome $\{\widetilde Y_i\}_{i=1}^N$.
  \begin{enumerate}[leftmargin=2.2em, itemsep=2pt, topsep=4pt]
    \item Partition $\{1, \dots, N\}$ into folds $\mathcal{F}_1, \dots,
      \mathcal{F}_K$. Let $\mathrm{fold}(i)$ denote the fold
      containing unit $i$.
    \item \textbf{Inner pass.} For $k = 1, \dots, K$:
      \begin{enumerate}[leftmargin=1.7em, itemsep=1pt, topsep=2pt]
        \item Fit $\hat\mu^{(-k)}$ on
          $\{(Z_i, A_i) : R_i V_i = 1,\; i \notin \mathcal{F}_k\}$.
        \item For audited units $i \in \mathcal{F}_k$ ($R_i = 1$), set
          $\widehat M_i = \hat\mu^{(-k)}(Z_i)
          + (V_i / \rho_i)\bigl(A_i - \hat\mu^{(-k)}(Z_i)\bigr)$.
      \end{enumerate}
    \item \textbf{Outer pass.} For $k = 1, \dots, K$:
      \begin{enumerate}[leftmargin=1.7em, itemsep=1pt, topsep=2pt]
        \item Fit $\hat g^{(-k)}$ on
          $\{(B_i, \widehat M_i) : R_i = 1,\; i \notin \mathcal{F}_k\}$,
          using the out-of-fold $\widehat M_i$ from step 2(b).
        \item For \emph{all} units $i \in \mathcal{F}_k$, set
          $\widetilde Y_i = \hat g^{(-k)}(B_i)
          + (R_i / \pi_i)\bigl(\widehat M_i - \hat g^{(-k)}(B_i)\bigr)$.
      \end{enumerate}
    \item \textbf{Return} $\{\widetilde Y_i\}_{i=1}^N$.
  \end{enumerate}
\end{algorithm}

\paragraph{Cross-fitting.}

Algorithm~\ref{alg:ngdsl} uses a shared fold partition for the inner and outer passes, which is the implementation used in the experiments. For the cleanest theoretical argument, Appendix \ref{appx:theory} first analyzes a leakage-free nested variant in which the outer nuisance for a held-out fold is trained using audit pseudo-labels constructed without any information from that fold. The shared-partition version differs only through an indirect dependence of the
outer nuisance on the held-out fold through the inner pseudo-labels used in outer training. Appendix \ref{appx:theory} isolates this difference as a stability remainder and states the high-level condition under which the shared-partition implementation has the same first-order expansion.

Write $m(W_i; \beta, \eta) = \X_i\,(\widetilde Y_i(\eta) - \expit(\X_i^\top \beta))$ for the estimating function, with $\eta = (\mu, g)$ now an explicit argument, and let $\beta_0$ denote the population value of $\beta$ defined
by \cref{eq:beta-pop}.

\begin{proposition}[Design validity and asymptotic normality]
  \label{prop:can}
  Suppose \crefrange{ass:audit}{ass:coders} hold, the design weights
  \(\pi_i,\rho_i\) are known and bounded away from zero and one,
  \(\E\|\X_i\|^{2+\delta}<\infty\) for some \(\delta>0\), and
  \[
    \mathbf H
    =
    \E\!\left[
      \X_i\X_i^\top \expit'\!\left(\X_i^\top\beta_0\right)
    \right]
  \]
  is nonsingular. For the leakage-free nested cross-fit estimator, under cross-fit
\(L^2\)-consistent nuisance estimation, \(\hat\beta\) defined by
\cref{eq:beta-emp} satisfies
  \[
    \sqrt N(\hat\beta-\beta_0)
    =
    \mathbf H^{-1}
    \frac{1}{\sqrt N}\sum_{i=1}^N
    m(W_i;\beta_0,\eta_0)
    + o_p(1),
  \]
  and hence
  \[
    \sqrt N(\hat\beta-\beta_0)
    \rightsquigarrow
    \mathcal N\!\left(
      \mathbf 0,\,
      \mathbf H^{-1}\Omega \mathbf H^{-1\top}
    \right),
  \]
  where
  \[
    \Omega
    =
    \E\!\left[
      m(W_i;\beta_0,\eta_0)m(W_i;\beta_0,\eta_0)^\top
    \right]
    =
    \operatorname{Var}\{m(W_i;\beta_0,\eta_0)\},
    \qquad
    \eta_0=(\mu_0,g_0).
  \]
  The equality to the variance holds because
  \(\E\{m(W_i;\beta_0,\eta_0)\}=\mathbf 0\). The covariance matrix
  \(\mathbf H^{-1}\Omega\mathbf H^{-1\top}\) is consistently estimated by the
  empirical cross-fit sandwich estimator. Moreover, for any fixed nuisance
  pair \(\eta=(\mu,g)\),
  \[
    \E\!\left[
      m(W_i;\beta_0,\eta)
    \right]
    =
    \mathbf 0,
  \]
  so the score is design-unbiased at \(\beta_0\) regardless of nuisance
  specification. The same expansion also holds for the shared-partition implementation in
\cref{alg:ngdsl} under the high-level stability condition stated in \cref{appx:theory}. Full assumptions and the influence-function
derivation for the leakage-free nested estimator are given in
\cref{appx:theory}.
\end{proposition}


Unlike doubly robust settings with estimated sampling weights, no product-rate
condition is needed because the known design weights make the linear nuisance
remainder conditionally mean-zero; Appendix~\ref{appx:theory} gives the full
argument and the shared-partition stability condition.

We instantiate \cref{alg:ngdsl} with \(K=5\) cross-fitting folds and use
\(L^2\)-regularized logistic regression for both \(\hat\mu\) and \(\hat g\).
In the experiments, we balance folds on the twenty \(Q\)-quintile
\(\times\) \(B_{\mathrm{diff}}\)-quartile cells as a finite-sample
stabilization heuristic. This fold stratification is not part of the
identifying design and is not required for design validity; ordinary random
folds or other balanced fold constructions could also be used. Prediction
clipping, the full estimator catalogue, and an ablation over the inner feature
set appear in \cref{appx:impl,appx:ablation}.

\section{Experimental Design}
\label{sec:experiments}

Because $\Ystar$ is latent in real annotation problems, empirical benchmarks must either assume a proxy truth or move to simulation. We therefore combine a synthetic Monte Carlo study, where $\Ystar$ is known by construction and the
measurement design can be varied, with a semi-synthetic study on the Wikipedia Detox corpus \citep{wulczyn2017detox}, where real text and crowdworker labels replace the parametric annotation model but the target is necessarily a reference-panel proxy. The two studies respectively test design validity against a known target and robustness under real annotator disagreement.

\paragraph{Synthetic data-generating process.}
Each replication generates $N=50{,}000$ units. The design is intentionally richer than independent label-flip noise. Motivated by work on human label variation and learning from disagreement, we allow some items to be intrinsically ambiguous, some coders to be systematically more or less reliable, and machines and humans to have partially distinct error modes \citep{plank2022problem,uma2021learning,jiang2022investigating}. Latent labels $Y_i^\star$ are drawn from a logistic model in $p=5$ correlated covariates. Three latent difficulty channels then modulate human and surrogate
accuracy: $D_i$, a shared item-difficulty channel affecting both humans and machines; $H_i$, a human-only nuisance channel; and $M_i$, a machine-only nuisance channel. The surrogate $Q_i$ depends on $Y_i^\star$, covariates, $D_i$,
and $M_i$; the coder labels $G_{i1},G_{i2}$ depend on $Y_i^\star$, coder-specific sensitivity/specificity, $D_i$, and $H_i$. Thus coder errors are heterogeneous and correlated through hard items, and surrogate errors are allowed to differ from human errors. This structure mirrors empirical annotation settings in which disagreement reflects a mixture of ambiguity, annotator effects, and task artifacts, while downstream regression remains vulnerable to
response misclassification \citep{neuhaus1999bias,wang2020methods}. The audit indicator is uniform Bernoulli($\pi = 0.10$), reflecting a modest
manual-review budget. Adjudication is either uniform at mean rate $\bar\rho$ or disagreement-weighted to mean $\bar\rho$. The latter targets cases where the two coders disagree, motivated by evidence that disagreement often reflects
item ambiguity, task subjectivity, or guideline instability rather than only idiosyncratic annotator noise
\citep{plank2022problem,uma2021learning,jiang2022investigating}. We summarize the simulation design with three primary scenarios: \emph{benign} (high-quality surrogate, low coder noise, uniform adjudication, $\bar\rho = 0.25$); \emph{realistic} (medium-quality surrogate, moderate coder noise, disagreement-driven adjudication, $\bar\rho = 0.25$); and \emph{hard} (weak surrogate, high coder heterogeneity, sparse adjudication, $\bar\rho = 0.10$). Full equations, default parameter values, and the scenario parameter table appear in \cref{appx:dgp}.

\paragraph{Semi-synthetic study (Wikipedia Detox).}
The Wikipedia Detox corpus contains talk-page comments labeled by crowdworkers for whether the comment contains a personal attack (i.e., hostile or insulting language directed at another contributor). In our notation, \(\Ystar_i\) denotes whether comment \(i\) would be judged to contain a personal attack under the target labeling standard. Because this status is unobserved in the Wikipedia data, the study is not a field validation against external ground truth. We instead evaluate against a researcher-constructed proxy truth estimated from a held-out reference panel of crowdworkers. Thus, the Detox study evaluates recovery of a reference-panel operationalization of the construct under real annotator disagreement, not validity against an external oracle for personal attack. In each replication, we re-draw the audit and adjudication samples under known design probabilities. The corpus comprises \(N \approx 1.16 \times 10^5\) talk-page comments with multiple crowdworker attack ratings per comment. Real-world covariates retained for the analysis are the editor's logged-in status, namespace indicator, comment year (standardized), and standardized log comment length; the surrogate \(Q_i\) is a TF--IDF logistic classifier fit out-of-fold on the comment text. We partition the crowdworkers into a reference panel (used for the proxy truth) and two audit panels (used as \(G_{i,1},G_{i,2}\)). The main-text proxy truth is a binary Dawid--Skene estimate from a held-out reference panel \citep{dawidskene1979}; results under hard-majority and soft-mean-share proxies appear in \cref{appx:detox-extra}. Audit and adjudication indicators are re-drawn each replication from \(R_i \sim \mathrm{Bernoulli}(0.10)\) and disagreement-weighted \(\rho_i\) with \(\bar\rho = 0.25\); we use \(R = 200\) replications.

\paragraph{Estimators.}
\Cref{tab:estimators} summarizes the main estimators. \PADSL{} uses
\cref{eq:Ytilde-expanded} with inner features
$Z_i=(Q_i,\X_i,G_{i,1},G_{i,2})$ and L2-regularized logistic nuisances.
Appendix \cref{appx:ablation} reports a variant that augments the inner
feature set with coder-pair indicators.

\begin{table}[t]
\centering
\caption{Estimators compared in the main experiments.}
\label{tab:estimators}
\begin{tabular}{lll}
\toprule
Group & Estimator & Description \\
\midrule
Plug-in & \textsc{Surrogate-only} & Plug in $Q_i$ \\
Plug-in & \textsc{Human-naive} & HT estimator using $G_{i,1}$ with weight $1/\pi$ \\
Noisy-gold DSL & \textsc{DSL-MajVote} & Standard DSL using two-coder majority label \\
Valid feasible & \textsc{DSL-AdjOnly} & DSL on adjudicated cases with weight $1/(\pi\rho_i)$ \\
Proposed & \PADSL{} & Nested audit/adjudication pseudo-outcome \\
Infeasible & \textsc{DSL-Oracle} & DSL using $\Ystar_i$ on the audit set \\
\bottomrule
\end{tabular}
\end{table}

\section{Results}
\label{sec:results}

We organize the empirical results around four claims, focusing on the logistic-regression slope $\beta_1$ throughout. The synthetic simulation analysis uses $R = 1{,}000$ replications per scenario with true $\beta_1 = 1$; the Detox study uses $R = 200$ replications with the Dawid--Skene proxy truth as the target. Prevalence and the inner-stage ablation are tabulated in \cref{appx:prevalence,appx:ablation}.

\begin{figure}[t]
  \centering
  \includegraphics[width=\linewidth]{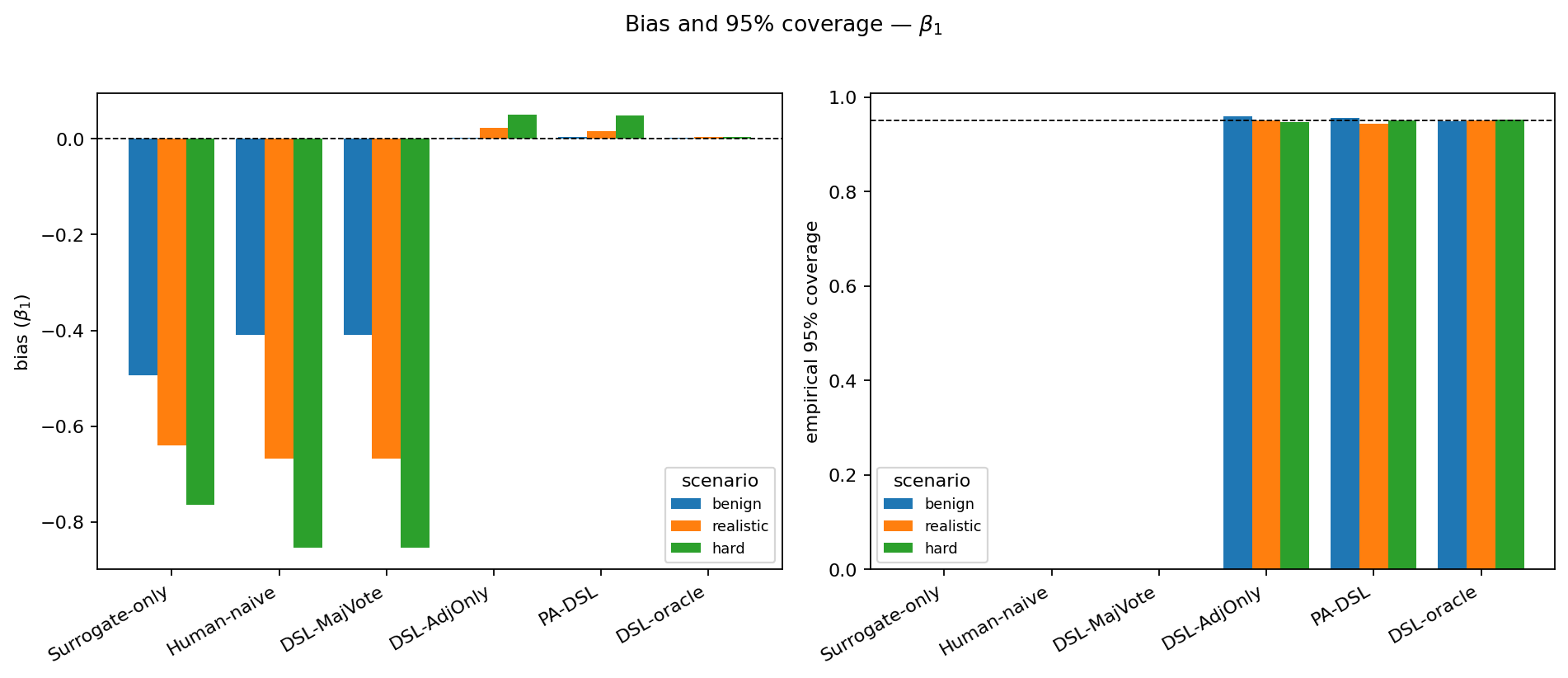}
  \caption{Synthetic study, $\beta_1$ across three scenarios ($R = 1{,}000$
    Monte Carlo replications, true $\beta_1 = 1$). \emph{Left:} bias.
    \emph{Right:} empirical $95\%$ coverage. The four plug-in baselines on
    the left of each panel are biased $-0.41$ to $-0.85$ with zero coverage;
    the three audit-corrected estimators on the right (\textsc{DSL-AdjOnly},
    \PADSL{}, \textsc{DSL-oracle}) recover near-zero bias and nominal
    coverage in every scenario.}
  \label{fig:beta1-bias-cov}
\end{figure}

\paragraph{Claim 1: noisy plug-ins fail.}
\Cref{fig:beta1-bias-cov} shows that \textsc{Surrogate-only}, \textsc{Human-naive}, and \textsc{DSL-MajVote} are all substantially biased, with bias ranging from $-0.41$ in the benign scenario to $-0.85$ in the hard scenario and empirical $95\%$ coverage equal to $0.000$ in every cell. Standard DSL does not repair this failure when its ``gold'' audit label is itself noisy: the correction removes surrogate-side bias relative to the audit label, and therefore inherits bias in that label. Majority vote also fails because coder errors are correlated through shared item difficulty, so the bias common to both coders survives aggregation.

\paragraph{Claim 2: audit-corrected estimators recover validity.}
\Cref{fig:beta1-bias-cov} also shows that \textsc{DSL-AdjOnly}, \PADSL{}, and the infeasible \textsc{DSL-oracle} all sit within Monte Carlo error of zero bias and report empirical coverage in $[0.94, 0.96]$ across all three scenarios. Adjudication-only DSL is design-valid because it uses only the adjudicated subset $\{R_i V_i = 1\}$ with the joint sampling weight $1/(\pi \rho_i).$ 
The price it pays is that the unadjudicated coder labels $G_{i,1}, G_{i,2}$ on the rest of the audit set are simply discarded.

\begin{figure}[t]
  \centering
  \includegraphics[width=\linewidth]{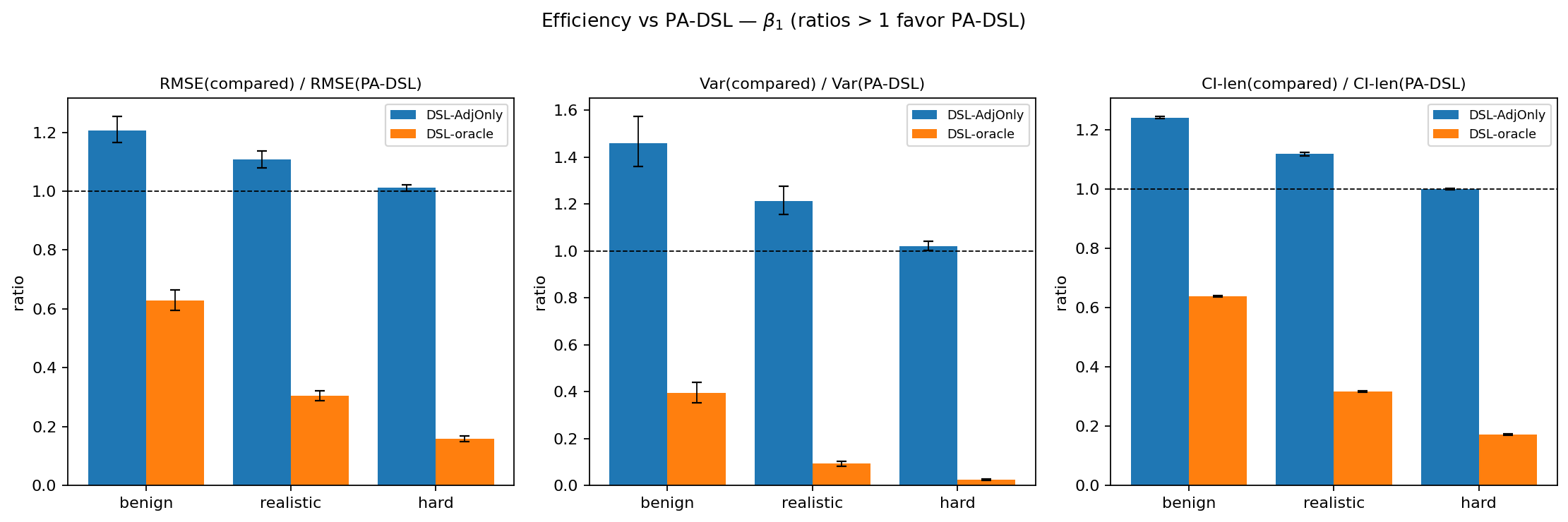}
  \caption{Relative efficiency of \PADSL{} on $\beta_1$. Bars above one favor
    \PADSL{}; whiskers are paired-bootstrap $95\%$ CIs over the $1{,}000$
    Monte Carlo replications. Against \textsc{DSL-AdjOnly} (blue),
    \PADSL{} achieves materially smaller variance and CI length in benign
    and realistic; in hard the inner stage extracts essentially no additional
    signal and the lower CI bound sits within $5 \times 10^{-4}$ of $1$.
    Against \textsc{DSL-oracle} (orange) the gap to the infeasible benchmark
    is roughly $1.6\times$ in benign and widens to $6.3\times$ in hard, the
    price paid for not observing $\Ystar$ on the audit set.}
  \label{fig:beta1-eff}
\end{figure}

\paragraph{Claim 3: \PADSL{} improves efficiency where the inner layer has signal.}
\Cref{fig:beta1-eff} reports the efficiency ratios of \PADSL{} versus \textsc{DSL-AdjOnly} (the strongest feasible baseline) and
\textsc{DSL-oracle} (the infeasible benchmark), with paired-bootstrap $95\%$ CIs over the $1{,}000$ Monte Carlo replications. Relative to \textsc{DSL-AdjOnly}, the RMSE ratio is $1.21$ ($95\%$ CI $[1.17, 1.25]$) in benign, $1.11$ ($[1.08, 1.14]$) in realistic, and $1.01$
($[1.00, 1.02]$) in hard, with corresponding variance ratios $1.46$ ($[1.36, 1.57]$), $1.21$ ($[1.16, 1.27]$), and $1.02$ ($[1.00, 1.04]$).
The benign and realistic gains are clear; the hard-regime ratio's lower bound sits within $5 \times 10^{-4}$ of $1$, so $\PADSL{}$ \emph{reverts to the adjudication-only benchmark} when the inner layer has no recoverable signal. In this scenario, the surrogate is weak, the coders are noisy, and the adjudication budget is sparse, so predicting \(A_i\) from \((Q_i,X_i,G_{i,1},G_{i,2})\) buys little leverage beyond the \(\pi_i\rho_i\)-weighted adjudicated subset. The reversion is the right behavior: \PADSL{} does not manufacture signal absent from the inner features. Classification-oriented diagnostics in Appendix \ref{app:classification-diag} show the same pattern at the prediction level: \(\widehat\mu(Z_i)\) improves on \(\widehat g(B_i)\) in the benign and realistic settings, but adds little signal in the hard setting.

\paragraph{Claim 4: Detox replicates the qualitative pattern.}
\Cref{tab:detox-beta1} shows the same comparison on the Wikipedia Detox corpus with the Dawid--Skene proxy truth ($\beta_1 = -1.42$ on \texttt{logged\_in}). The plug-in baselines and the standard-DSL variants fed with noisy gold are biased by $0.33$--$0.39$ with zero empirical coverage; \textsc{DSL-AdjOnly} and \PADSL{} both achieve near-zero bias and coverage at or slightly above $95\%$. \PADSL{}'s RMSE ratio against \textsc{DSL-AdjOnly} is $1.18$ ($95\%$ paired-bootstrap CI $[1.10, 1.27]$) and its CI-length ratio is $1.13$ ($[1.12, 1.13]$); both are clearly above $1$, so the efficiency gain is statistically distinguishable from the no-gain point. The same qualitative ranking is reproduced under the hard-majority and soft-mean proxies in \cref{appx:detox-extra}. As above, this is evidence for the sampling-and-adjudication machinery under real text and real crowdworker disagreement, with respect to the constructed proxy truth rather than an observed latent construct.

\begin{table}[t]
  \caption{Detox semi-synthetic study, $\beta_1$ on \texttt{logged\_in} under
    the Dawid--Skene proxy truth ($R = 200$ replications, target
    $\beta_1 = -1.42$). \textbf{Bold} marks the column-best for MC SD,
    RMSE, and CI length among design-valid estimators (\textsc{DSL-AdjOnly},
    \PADSL{}); coverage is a calibration diagnostic and bias is reported
    but not bolded. The light rule separates the infeasible
    \textsc{DSL-oracle} benchmark.}
  \label{tab:detox-beta1}
  \centering
  \small
  \begin{tabular}{lrrrrr}
    \toprule
    Method & Bias & MC SD & RMSE & Coverage & CI len. \\
    \midrule
    \textsc{Surrogate-only}     & $\phantom{-}0.387$ & $0.000$ & $0.387$ & $0.000$ & $0.043$ \\
    \textsc{Human-naive}        & $\phantom{-}0.332$ & $0.045$ & $0.335$ & $0.000$ & $0.208$ \\
    \textsc{DSL-MajVote}        & $\phantom{-}0.334$ & $0.037$ & $0.336$ & $0.000$ & $0.179$ \\
    \textsc{DSL-AdjOnly}        & $\phantom{-}0.005$ & $0.098$ & $0.098$ & $0.935$ & $0.368$ \\
    \PADSL{}                    & $\phantom{-}0.009$ & $\best{0.082}$ & $\best{0.083}$ & $0.975$ & $\best{0.327}$ \\
    \oraclesep
    \textsc{DSL-oracle}         & $\phantom{-}0.005$ & $0.035$ & $0.036$ & $0.965$ & $0.144$ \\
    \bottomrule
  \end{tabular}
\end{table}

\section{Discussion}
\label{sec:discussion}

\PADSL{} extends design-based supervised learning to hierarchical annotation
pipelines in which adjudication is treated as the gold-standard label but is
available only for a subset of audited cases. The results suggest several
practical lessons for designing and analyzing such pipelines.

\paragraph{When the proposal helps and when it does not.}
\PADSL{} delivers efficiency gains over adjudication-only DSL when the audit-tier
features \(Z_i=(Q_i,X_i,G_{i,1},G_{i,2})\) predict the adjudicated label \(A_i\)
beyond the outer features \(B_i=(Q_i,X_i)\). When this inner layer contains little
additional signal, \PADSL{} behaves like the adjudication-only benchmark, as in
the hard synthetic regime. This pattern is formalized by the variance comparison
in \cref{appx:efficiency-gain}, where the gain over adjudication-only DSL is
proportional to the \(Z\)-explainable component \(\mu_0(Z_i)-g_0(B_i)\). The
inner ablations in \cref{appx:inner-ablation} also show that the augmented-IPW
form is not merely cosmetic: regression-only correction produced severe
undercoverage in the realistic and hard regimes, underscoring the need to
combine the outcome model with the known adjudication design.

\paragraph{Design implications.}
The design sensitivity analysis in \cref{appx:design-curves} suggests a
practical operating range in the settings we studied: audit rates of roughly
\(\pi \in [0.10,0.20]\) combined with disagreement-driven adjudication at
\(\bar\rho \in [0.10,0.25]\). Disagreement-driven adjudication should include a
positive adjudication floor even among agreement cases; otherwise, the design
cannot identify error patterns in apparently easy items. More generally, the
two-coder setup used in our experiments is not required by the framework:
\(Z_i\) may include any audit-tier information observed when \(R_i=1\),
including \(K_i\) coder labels, coder identities, agreement indicators, vote
shares, uncertainty scores, or other metadata. The same inner correction applies
as long as the adjudication probability
\(\rho_i=P(V_i=1\mid R_i=1,Z_i)\) is known and bounded away from zero. The realized audit and adjudication probabilities should also be recorded with the labels, since
\PADSL{} relies on known \(\pi_i\) and \(\rho_i\) rather than estimating them
after the fact. A pilot audit can help assess whether coder labels contain
enough signal about the adjudicated construct to justify collecting additional
human labels before increasing the adjudication budget.

\paragraph{Scope and limitations.}
Although LLM-based annotation is one motivating use case, PA-DSL is a classifier-agnostic design-based correction: the surrogate \(Q_i\) may come from an LLM, a supervised classifier, a rules-based score, or any other automated labeling system observed on the full frame. However, \PADSL{} is intended for prospective, known-probability audit/adjudication designs, or for settings in which the audit and adjudication probabilities can be reconstructed from the design record. It is not a generic post hoc correction for arbitrary deterministic review pipelines: if audited or adjudicated cases were selected using undocumented rules, or if some design strata have zero adjudication probability, the required positivity and ignorability conditions may fail.

Validity also relies on \cref{ass:correct}. We interpret this assumption as requiring adjudication to recover the target label under the adjudication protocol. \Cref{appx:imperfect-adj} shows that \PADSL{} inherits adjudicator bias relative to $\Ystar_i$ when \cref{ass:correct} fails. Thus, \PADSL{} corrects the sampling and measurement structure only relative to the
adjudicated construct; it cannot by itself guarantee that the adjudication protocol captures the substantive construct researchers intended. Estimators that remain valid under imperfect adjudication, treating $A_i$ as a noisy proxy with a known design weight, are a natural next step.

The Detox study should be interpreted in this same spirit. It is a semi-synthetic evaluation using real text and real crowdworker disagreement, but the audit/adjudication indicators are re-drawn with known probabilities and the target is a held-out reference-panel operationalization rather than an external field-validated oracle. Non-binary outcomes, scores nonlinear in $\Ystar_i$, and dependent-class missingness lie outside the current scope.


\bibliographystyle{plainnat}
\bibliography{refs}

\appendix
\section{Prevalence as a Special Case}
\label{appx:prevalence}

The population prevalence $\psi = \E[\Ystar]$ is the special case of the
outcome-affine framework in which the score is $\Ystar_i - \psi$. The
\PADSL{} estimator instantiates as
$\hat\psi = N^{-1} \sum_i \widetilde Y_i$ with $\widetilde Y_i$ from
\cref{eq:Ytilde-expanded}, and the empirical-variance standard error
$\widehat{\mathrm{SE}}(\hat\psi) = (N(N-1))^{-1/2}
\bigl(\sum_i (\widetilde Y_i - \hat\psi)^2\bigr)^{1/2}$ is consistent for
$N^{-1/2}$ times the standard deviation of the influence function in
\cref{eq:if}. \Cref{tab:appx-prev-syn} reports the synthetic results for
this target across the three scenarios; the qualitative ranking matches the
$\beta_1$ results in the main text.

\begin{table}[h]
  \caption{Synthetic study, prevalence $\psi$ ($R = 1{,}000$ replications,
    realized truth $\bar\Ystar \approx 0.426$). \textbf{Bold} marks the
    column-best for MC SD, RMSE, and CI length per scenario among
    design-valid estimators (\textsc{DSL-AdjOnly}, \PADSL{}); coverage is
    a calibration diagnostic and bias is reported but not bolded. The
    light rule separates the infeasible \textsc{DSL-oracle} benchmark.}
  \label{tab:appx-prev-syn}
  \centering
  \small
  \begin{tabular}{llrrrrr}
    \toprule
    Scenario & Method & Bias & MC SD & RMSE & Coverage & CI len. \\
    \midrule
    benign & \textsc{Surrogate-only}    & $\phantom{-}0.0905$ & $0.0012$ & $0.0905$ & $0.000$ & $0.0047$ \\
           & \textsc{Human-naive}       & $\phantom{-}0.0010$ & $0.0110$ & $0.0111$ & $0.898$ & $0.0354$ \\
           & \textsc{DSL-MajVote}       & $\phantom{-}0.0008$ & $0.0085$ & $0.0086$ & $0.817$ & $0.0232$ \\
           & \textsc{DSL-AdjOnly}       & $\phantom{-}0.0003$ & $0.0078$ & $0.0078$ & $0.945$ & $0.0305$ \\
           & \PADSL{}                   & $\phantom{-}0.0003$ & $\best{0.0063}$ & $\best{0.0063}$ & $0.949$ & $\best{0.0247}$ \\
    \oraclesep
           & \textsc{DSL-oracle}        & $-0.0000$           & $0.0042$ & $0.0042$ & $0.948$ & $0.0163$ \\
    \midrule
    realistic & \textsc{Surrogate-only} & $-0.0605$           & $0.0012$ & $0.0605$ & $0.000$ & $0.0049$ \\
           & \textsc{Human-naive}       & $\phantom{-}0.0014$ & $0.0173$ & $0.0174$ & $0.692$ & $0.0354$ \\
           & \textsc{DSL-MajVote}       & $\phantom{-}0.0011$ & $0.0162$ & $0.0162$ & $0.570$ & $0.0266$ \\
           & \textsc{DSL-AdjOnly}       & $-0.0002$           & $0.0193$ & $0.0193$ & $0.941$ & $0.0750$ \\
           & \PADSL{}                   & $-0.0003$           & $\best{0.0170}$ & $\best{0.0170}$ & $0.951$ & $\best{0.0669}$ \\
    \oraclesep
           & \textsc{DSL-oracle}        & $-0.0000$           & $0.0056$ & $0.0056$ & $0.949$ & $0.0220$ \\
    \midrule
    hard   & \textsc{Surrogate-only}    & $-0.1434$           & $0.0013$ & $0.1435$ & $0.000$ & $0.0051$ \\
           & \textsc{Human-naive}       & $\phantom{-}0.0074$ & $0.0223$ & $0.0235$ & $0.543$ & $0.0357$ \\
           & \textsc{DSL-MajVote}       & $\phantom{-}0.0074$ & $0.0219$ & $0.0231$ & $0.434$ & $0.0272$ \\
           & \textsc{DSL-AdjOnly}       & $-0.0004$           & $0.0341$ & $0.0341$ & $0.938$ & $0.1294$ \\
           & \PADSL{}                   & $-0.0003$           & $\best{0.0339}$ & $\best{0.0338}$ & $0.938$ & $\best{0.1293}$ \\
    \oraclesep
           & \textsc{DSL-oracle}        & $\phantom{-}0.0000$ & $0.0060$ & $0.0060$ & $0.948$ & $0.0237$ \\
    \bottomrule
  \end{tabular}
\end{table}
\section{Identification, Influence Function, and Variance}
\label{appx:theory}

\paragraph{Notation and population nuisances.}
For notational simplicity, this appendix writes the audit and adjudication probabilities as \(\pi(B_i)\) and \(\rho(Z_i)\). If the actual audit probability depends on additional observed design variables, those variables
are either included in \(B_i\) or added to the conditioning set below. Likewise, any observed variables used to set adjudication probabilities are included in \(Z_i\). The arguments require only that the relevant conditional
inclusion probabilities are known by design and bounded away from zero.

We write the population nuisance functions as
\[
  \mu_0(z)=\E(A_i\mid Z_i=z,R_i=1,V_i=1),
  \qquad
  g_0(b)=\E\{\mu_0(Z_i)\mid B_i=b,R_i=1\}.
\]
Under Assumptions~1--3,
\(\mu_0(Z_i)=\E(Y_i^\star\mid Z_i,R_i=1)\) and
\(g_0(B_i)=\E(Y_i^\star\mid B_i)\). Thus \(g_0\) is a population target; it does not depend on the realized nuisance estimate or on a particular cross-fit pseudo-label.

\paragraph{Pointwise identification.}
Under \cref{ass:audit}, $\E[\Ystar_i \mid \X_i, Q_i]
= \E[\Ystar_i \mid \X_i, Q_i, R_i = 1]$. Under
\cref{ass:adj,ass:correct},
$\E[\Ystar_i \mid Z_i, R_i = 1] = \E[A_i \mid Z_i, R_i V_i = 1] = \mu_0(Z_i)$.
Composing,
$\E[\mu_0(Z_i) \mid B_i, R_i = 1] = \E[\Ystar_i \mid B_i, R_i = 1]
= \E[\Ystar_i \mid B_i] = g_0(B_i)$.

\paragraph{Design-validity of $\tilde Y_i$.} Fix arbitrary nuisance functions
$\eta=(\mu,g)$. For audited units define
\[
M_i(\mu)=\mu(Z_i)+\frac{V_i}{\rho(Z_i)}\{A_i-\mu(Z_i)\}.
\]
When the nuisances are estimated by cross-fitting, the following argument is
understood conditionally on the training folds, so that $\mu$ and $g$ are fixed
functions for the evaluation fold.

First consider the inner correction. Conditional on $R_i=1$ and $Z_i$,
\[
\begin{aligned}
E\{M_i(\mu)\mid R_i=1,Z_i\}
&=
\mu(Z_i)+
E\left[
\frac{V_i}{\rho(Z_i)}\{A_i-\mu(Z_i)\}
\mid R_i=1,Z_i
\right] \\
&=
\mu(Z_i)+
E\{A_i-\mu(Z_i)\mid R_i=1,Z_i,V_i=1\} \\
&=
E(A_i\mid R_i=1,Z_i,V_i=1).
\end{aligned}
\]
By adjudication correctness, $A_i=Y_i^\star$ on $\{R_iV_i=1\}$, and by
adjudication ignorability,
\[
E(A_i\mid R_i=1,Z_i,V_i=1)
=
E(Y_i^\star\mid R_i=1,Z_i).
\]
Hence
\[
E\{M_i(\mu)\mid R_i=1,Z_i\}
=
E(Y_i^\star\mid R_i=1,Z_i).
\]

No restriction on the joint distribution of $(G_{i,1},G_{i,2})$ is used here: once we condition on $Z_i$, the coder labels are simply observed covariates, and all coder dependence or bias is part of the conditional mean $\mu_0(Z_i)$.

Now consider the outer correction. Conditional on $B_i$,
\[
\begin{aligned}
E\{\tilde Y_i(\eta)\mid B_i\}
&=
g(B_i)+
E\left[
\frac{R_i}{\pi(B_i)}
\{M_i(\mu)-g(B_i)\}
\mid B_i
\right] \\
&=
g(B_i)+
E\{M_i(\mu)-g(B_i)\mid R_i=1,B_i\} \\
&=
E\{M_i(\mu)\mid R_i=1,B_i\}.
\end{aligned}
\]
Using iterated expectations and the inner result,
\[
\begin{aligned}
E\{M_i(\mu)\mid R_i=1,B_i\}
&=
E\left[
E\{M_i(\mu)\mid R_i=1,Z_i\}
\mid R_i=1,B_i
\right] \\
&=
E\left[
E(Y_i^\star\mid R_i=1,Z_i)
\mid R_i=1,B_i
\right] \\
&=
E(Y_i^\star\mid R_i=1,B_i).
\end{aligned}
\]
Finally, by audit ignorability,
\[
E(Y_i^\star\mid R_i=1,B_i)=E(Y_i^\star\mid B_i).
\]
Therefore
\[
E\{\tilde Y_i(\eta)\mid B_i\}
=
E(Y_i^\star\mid B_i).
\]
Since $X_i$ is measurable with respect to $B_i$,
\[
E\left[
X_i\{\tilde Y_i(\eta)-\operatorname{expit}(X_i^\top\beta)\}
\right]
=
E\left[
X_i\{Y_i^\star-\operatorname{expit}(X_i^\top\beta)\}
\right].
\]
At $\beta=\beta_0$, the right-hand side is zero by definition of $\beta_0$. Thus $E\{m(W_i;\beta_0,\eta)\}=0$ for every fixed nuisance pair $\eta$. This fixed-nuisance statement establishes design-unbiasedness of the score, but it does not imply that arbitrary inconsistent nuisance estimators have the same first-order expansion. The influence-function representation below is obtained by expanding around $\eta_0=(\mu_0,g_0)$ and therefore requires the stated cross-fit nuisance convergence conditions.

\paragraph{Influence function.}
The cross-fit estimator $\hat\psi$ admits the asymptotic linear expansion
$\sqrt N (\hat\psi - \psi) = N^{-1/2} \sum_i \varphi(W_i; \mu_0, g_0) + o_p(1)$
with influence function
\begin{equation}
  \varphi(W_i; \mu_0, g_0)
  = g_0(B_i) - \psi
  + \frac{R_i}{\pi_i}\bigl[\mu_0(Z_i) - g_0(B_i)\bigr]
  + \frac{R_i V_i}{\pi_i \rho_i}\bigl[A_i - \mu_0(Z_i)\bigr].
  \label{eq:if}
\end{equation}
The estimator $\hat\beta$ from \cref{eq:beta-emp} has influence function
\begin{equation}
  \mathbf{H}^{-1}\,\X_i\,\bigl(\widetilde Y_i(\eta_0) - \expit(\X_i^\top \beta_0)\bigr),
  \qquad
  \mathbf{H} = \E[\X_i \X_i^\top\,\expit'(\X_i^\top \beta_0)],
\end{equation}
obtained directly from the M-estimator score $m(W_i; \beta_0, \eta_0)$;
the prevalence-target $\varphi$ is not the right object for $\hat\beta$
because the logistic score is not the prevalence influence function.

\paragraph{Why no product-rate condition is required.}
Let $\delta_g(B) = \hat g(B) - g_0(B)$ and
$\delta_\mu(Z) = \hat\mu(Z) - \mu_0(Z)$. Direct expansion of
$\widetilde Y_i$ gives
\begin{equation}
  \widetilde Y_i(\hat\eta) - \widetilde Y_i(\eta_0)
  =
  \Bigl(1 - \frac{R_i}{\pi_i}\Bigr)\delta_g(B_i)
  +
  \frac{R_i}{\pi_i}
  \Bigl(1 - \frac{V_i}{\rho_i}\Bigr)\delta_\mu(Z_i),
  \label{eq:nuis-perturb}
\end{equation}
so that
$m(W_i;\beta_0,\hat\eta) - m(W_i;\beta_0,\eta_0)
= \X_i \cdot [\text{r.h.s.\ of \cref{eq:nuis-perturb}}]$.
The two nuisance errors enter linearly, but each is multiplied by a
known-design, conditionally mean-zero sampling residual. This is why the
argument requires only $L^2$ consistency of the nuisance estimators, rather
than a product-rate condition.

\paragraph{Leakage-free nested cross-fitting.}
We first analyze a leakage-free nested version of the estimator. For a held-out
evaluation fold $\mathcal F_k$, the inner nuisance used to construct audit
pseudo-labels for the outer training sample is fit without using observations
from $\mathcal F_k$. The outer nuisance used to evaluate $\mathcal F_k$ is
therefore trained on pseudo-labels that contain no information from the held-out
fold. Conditional on the training data for that fold, the nuisance functions are
fixed with respect to the evaluation observations.

For such an evaluation observation, the conditional means of both bracketed
factors in \cref{eq:nuis-perturb} are zero:
\[
  \E\!\left[1 - \frac{R_i}{\pi_i} \,\middle|\, B_i\right] = 0
\]
by \cref{ass:audit}, and
\[
  \E\!\left[
    \frac{R_i}{\pi_i}
    \Bigl(1 - \frac{V_i}{\rho_i}\Bigr)
    \,\middle|\, Z_i,B_i
  \right] = 0
\]
by \cref{ass:adj}; the inner factor has conditional mean zero given
$R_i=1$ and $Z_i$. Hence
\begin{equation}
  \E\!\Bigl[
    m(W_i;\beta_0,\hat\eta) - m(W_i;\beta_0,\eta_0)
    \,\Big|\, \text{training data}
  \Bigr]
  =
  \mathbf 0.
  \label{eq:cond-zero-rem}
\end{equation}
Thus the nuisance error contributes only a centered empirical-process
remainder.

Because $\pi_i$ and $\rho_i$ are bounded away from zero, and
$\E\|\X_i\|^{2+\delta}<\infty$, there is a finite constant $C$ such that
\begin{equation}
  \E\!\Bigl[
    \bigl\|
      m(W_i;\beta_0,\hat\eta) - m(W_i;\beta_0,\eta_0)
    \bigr\|^2
    \,\Big|\, \text{training data}
  \Bigr]
  \leq
  C\left(
    \|\hat g-g_0\|_{2,X,B}^2
    +
    \|\hat\mu-\mu_0\|_{2,X,Z}^2
  \right),
  \label{eq:l2-rem-bound}
\end{equation}
where
\[
  \|h\|_{2,X,B}^2
  =
  \E\!\left[\|\X_i\|^2 h(B_i)^2\right],
  \qquad
  \|r\|_{2,X,Z}^2
  =
  \E\!\left[\|\X_i\|^2 r(Z_i)^2\right].
\]
Therefore,
\[
  \sqrt N(P_N-P)
  \{m(W;\beta_0,\hat\eta)-m(W;\beta_0,\eta_0)\}
  =
  O_p\!\left(
    \|\hat g-g_0\|_{2,X,B}
    +
    \|\hat\mu-\mu_0\|_{2,X,Z}
  \right)
  =
  o_p(1),
\]
under cross-fit $X$-weighted $L^2$-consistent nuisance estimation.

Combining this centered
empirical-process bound with the fixed-nuisance design-unbiasedness result
above and a standard mean-value expansion in $\beta$ around $\beta_0$ gives the
asymptotic linear expansion in \cref{prop:can} for the leakage-free nested
estimator. No product-rate condition appears because the known design weights
make the linear nuisance remainder conditionally mean-zero.

\paragraph{Shared-partition cross-fitting.}
Algorithm~\ref{alg:ngdsl} uses a computationally simpler shared-partition
implementation, in which the same folds are used in the inner and outer
passes. For fold $\mathcal F_k$, write $\hat g_{\mathrm{sp}}^{(-k)}$ for the
outer nuisance actually used by the algorithm. Let
$\hat g_{\mathrm{lf}}^{(-k)}$ denote the leakage-free counterpart: the outer
nuisance that would be obtained by constructing each training pseudo-label
$\widehat M_j$, $j\notin\mathcal F_k$, using an inner fit that excludes both
$\mathcal F_k$ and $\mathcal F_{\mathrm{fold}(j)}$.

The shared-partition estimator has the same first-order expansion as the
leakage-free estimator provided the leakage remainder satisfies
\begin{equation}
\frac{1}{\sqrt N}\sum_{k=1}^{K}\sum_{i \in \mathcal F_k}
X_i\Bigl(1 - \frac{R_i}{\pi_i}\Bigr)
\bigl\{
  \hat g_{\mathrm{sp}}^{(-k)}(B_i)
  - \hat g_{\mathrm{lf}}^{(-k)}(B_i)
\bigr\}
= o_p(1).
\tag{SP}
\label{eq:sp}
\end{equation}
To see this, add and subtract the leakage-free score:
\begin{align}
&\frac{1}{\sqrt N} \sum_{k=1}^{K} \sum_{i \in \mathcal F_k}
\left[
  m\bigl(W_i;\beta_0,\hat\mu^{(-k)},\hat g_{\mathrm{sp}}^{(-k)}\bigr)
  - m(W_i;\beta_0,\eta_0)
\right]
\notag\\
&\quad=
\frac{1}{\sqrt N} \sum_{k=1}^{K} \sum_{i \in \mathcal F_k}
\left[
  m\bigl(W_i;\beta_0,\hat\mu^{(-k)},\hat g_{\mathrm{lf}}^{(-k)}\bigr)
  - m(W_i;\beta_0,\eta_0)
\right]
\notag\\
&\qquad+
\frac{1}{\sqrt N}\sum_{k=1}^{K}\sum_{i \in \mathcal F_k}
X_i\Bigl(1 - \frac{R_i}{\pi_i}\Bigr)
\bigl\{
  \hat g_{\mathrm{sp}}^{(-k)}(B_i)
  - \hat g_{\mathrm{lf}}^{(-k)}(B_i)
\bigr\}.
\label{eq:sp-decomp}
\end{align}
The first term is $o_p(1)$ by the leakage-free argument. The second term is
$o_p(1)$ by condition~\eqref{eq:sp}. Hence the shared-partition estimator has
the same asymptotic linear representation as the leakage-free estimator
whenever condition~\eqref{eq:sp} holds.

\paragraph{Discussion of the shared-partition stability condition.}
Condition~\eqref{eq:sp} is a stability condition on the indirect leakage
created by reusing the same fold partition in the inner and outer stages.
It is expected to hold when the inner learner is stable under adding or
removing one fold, the outer learner is Lipschitz-stable in its training
responses, and the resulting difference
\[
  \|\hat g_{\mathrm{sp}}^{(-k)}-\hat g_{\mathrm{lf}}^{(-k)}\|_{2,X}
  = o_p(1)
\]
is sufficiently small that the weighted empirical average in
\eqref{eq:sp} is \(o_p(1)\). This is plausible for stable nuisance learners
in fixed-dimensional smooth classes, such as the fixed-hyperparameter
\(L^2\)-regularized logistic regressions used in the experiments, under
standard nonsingularity and bounded-moment conditions. A formal verification
requires learner-specific deletion stability controlling the effect of
excluding \(\mathcal F_k\) from the inner fits on the resulting outer
nuisance. The condition may fail for highly adaptive nuisance procedures,
including tuning, stacking, or model selection performed inside the
cross-fitting loop, because such procedures can transmit evaluation-fold
information into \(\hat g_{\mathrm{sp}}^{(-k)}\).


\paragraph{Variance decomposition.}
Let
\begin{equation}
  H(Z, B) \;=\; \mu_0(Z) - g_0(B),
  \qquad
  \varepsilon \;=\; A - \mu_0(Z),
\end{equation}
so that $\varphi(W; \mu_0, g_0) = \bigl(g_0(B) - \psi\bigr)
+ \tfrac{R}{\pi(B)} H(Z,B)
+ \tfrac{R V}{\pi(B)\,\rho(Z)} \varepsilon$.
Because $\E[H(Z, B) \mid B, R = 1] = 0$ (by definition of $g_0$) and
$\E[\varepsilon \mid Z, R = 1, V = 1] = 0$ (by definition of $\mu_0$), the
three summands are mutually orthogonal martingale increments. Iterated
expectations give
$\E[(R/\pi(B))^2 H(Z,B)^2]
= \E[H(Z,B)^2 / \pi(B)]$ and
$\E[(R V/(\pi(B)\rho(Z)))^2 \varepsilon^2]
= \E[\sigma_A^2(Z) / (\pi(B)\,\rho(Z))]$
where $\sigma_A^2(Z) = \mathrm{Var}(A \mid Z, R = 1, V = 1)$, so
\begin{equation}
  \mathrm{Var}[\varphi]
  \;=\; \mathrm{Var}[g_0(B)]
  \;+\; \E\!\left[\frac{H(Z, B)^2}{\pi(B)}\right]
  \;+\; \E\!\left[\frac{\sigma_A^2(Z)}{\pi(B)\,\rho(Z)}\right].
  \label{eq:var-ng}
\end{equation}
For constant $\pi(B) \equiv \pi$ and $\rho(Z) \equiv \rho$,
$\mathrm{Var}[\varphi] = \mathrm{Var}[g_0(B)]
+ \pi^{-1} \E[(\mu_0(Z) - g_0(B))^2]
+ (\pi \rho)^{-1} \E[\sigma_A^2(Z)]$.
At $\pi = \rho = 1$ this collapses to $\mathrm{Var}(A)$ via the law of total
variance, as it must.

\paragraph{Efficiency gain over \textsc{DSL-AdjOnly}.}
\label{appx:efficiency-gain}
The adjudication-only DSL pseudo-outcome treats $R_i V_i$ as the sampling
indicator and uses the joint design weight $1/(\pi(B_i)\,\rho(Z_i))$ to
debias the outer regression $\hat g$, giving influence function
$\varphi_{\mathrm{adj}} = (g_0(B) - \psi) + (R V/(\pi(B)\,\rho(Z)))(A - g_0(B))$
--- structurally a single-stage AIPW with the joint sampling indicator.
Decomposing $A - g_0(B) = H(Z, B) + \varepsilon$ and applying the same
orthogonality argument,
\begin{equation}
  \mathrm{Var}[\varphi_{\mathrm{adj}}]
  \;=\; \mathrm{Var}[g_0(B)]
  \;+\; \E\!\left[\frac{H(Z, B)^2}{\pi(B)\,\rho(Z)}\right]
  \;+\; \E\!\left[\frac{\sigma_A^2(Z)}{\pi(B)\,\rho(Z)}\right].
  \label{eq:var-adj}
\end{equation}
Subtracting \cref{eq:var-ng} from \cref{eq:var-adj} gives
\begin{equation}
  \mathrm{Var}[\varphi_{\mathrm{adj}}] - \mathrm{Var}[\varphi]
  \;=\; \E\!\left[
    \frac{1 - \rho(Z)}{\pi(B)\,\rho(Z)}\,
    \bigl(\mu_0(Z) - g_0(B)\bigr)^2
  \right] \;\geq\; 0,
  \label{eq:var-gain}
\end{equation}
which simplifies to $((1-\rho)/(\pi\rho))\,\E[(\mu_0(Z) - g_0(B))^2]$ for constant $\pi, \rho$. 

For the logistic-regression target in the main text, the same comparison applies
to the covariance of the score influence function
\(X_i\{\widetilde Y_i(\eta_0)-\expit(X_i^\top\beta_0)\}\). In particular, the
corresponding difference in score covariance matrices is
\begin{equation}
  \Omega_{\mathrm{adj}} - \Omega
  \;=\;
  \E\!\left[
    X_i X_i^\top
    \frac{1 - \rho(Z_i)}{\pi(B_i)\,\rho(Z_i)}\,
    \bigl(\mu_0(Z_i) - g_0(B_i)\bigr)^2
  \right],
  \label{eq:omega-gain}
\end{equation}
which is positive semidefinite. Applying the usual M-estimation transformation, the corresponding asymptotic covariance difference for \(\hat\beta\) is
\[
  H^{-1}(\Omega_{\mathrm{adj}}-\Omega)H^{-1\top},
\]
also positive semidefinite. Thus the scalar comparison in
\cref{eq:var-gain} has the same interpretation for the main
logistic-regression estimand: \PADSL{} removes the extra
\(\rho(Z_i)^{-1}\) inflation from the component of adjudicated truth explained
by audit-tier features beyond \(B_i\).

The \PADSL{} estimator does \emph{not} reduce the residual adjudication noise term $\sigma_A^2(Z)$, which can only be learned from adjudicated units and remains weighted by $1/(\pi(B)\rho(Z))$ in both estimators. What \PADSL{} does is move the $Z$-explainable component $\mu_0(Z) - g_0(B)$ from the sparse adjudication weight $1/(\pi(B)\rho(Z))$ to the larger audit weight $1/\pi(B)$. The gain in \cref{eq:var-gain} is therefore large precisely when (i) adjudication is sparse, so \(\rho(Z)\) is small, and (ii) the audit-level features $Z$ explain substantial variation in the adjudicated label, and collapses to zero when $Z$ has no incremental predictive power over $B$.

\section{Why build on DSL rather than PPI?}
\label{appx:ppi}

Prediction Powered Inference (PPI) is a closely related prediction-assisted inference framework \citep{angelopoulos2023prediction}. We build on DSL because our measurement process is explicitly design-based: the audit and adjudication samples are drawn with known probabilities chosen by the analyst. The DSL pseudo-outcome is therefore the natural object to extend, and partial adjudication is handled by nesting a second known-weight correction inside the usual audit correction. This
is not a claim that PPI is inappropriate; rather, deriving a partially adjudicated PPI analogue is a separate extension. Recent benchmarking finds that DSL can be competitive with, and often more efficient than, PPI for
LLM-based parameter estimation, though performance varies by setting \citep{de2025benchmarking}.

\section{Implementation Details}
\label{appx:impl}

All nuisances are L2-regularized logistic regression unless otherwise noted
($C = 1$, max-iter $2000$); predictions are clipped to
$[10^{-3}, 1 - 10^{-3}]$ to bound the importance weights in
\cref{eq:Ytilde-expanded}. Cross-fitting uses $K = 5$ folds, stratified on
twenty $Q$-quintile $\times$ $B_{\mathrm{diff}}$-quartile strata where
$B_{\mathrm{diff}} = 1 - 2|p^\star - 0.5|$ is the boundary statistic.
$\beta$ is fit by damped iteratively reweighted least squares with a
$10^{-6}$ ridge for numerical stability; standard errors use the empirical
sandwich on the cross-fit pseudo-outcome.

The estimators evaluated in the main text and appendices are:
\textsc{Surrogate-only}, \textsc{Human-naive},
\textsc{DSL-MajVote} (two coders, ties randomly broken),
\textsc{DSL-AdjOnly}, \textsc{DSL-oracle},
\PADSL{} (logistic, baseline inner features),
\textsc{PA-DSL-coder} (logistic, coder-pair features),
\textsc{PA-DSL-IPW-only} (inner $\hat\mu \equiv 0$; pure Horvitz--Thompson
at the inner stage), and \textsc{PA-DSL-Reg-only} (no $V/\rho$ debias at
the inner stage; pure regression).

\section{Synthetic Data-Generating Process}
\label{appx:dgp}

Each replication generates $N = 50{,}000$ units. Covariates $\X_i$ comprise
an intercept and $p = 5$ continuous features drawn from a multivariate
normal with unit variances and equicorrelation $\rho_{\mathrm{cov}} = 0.2$.
Latent labels are
$\Ystar_i \mid \X_i \sim \text{Bernoulli}(\expit(\X_i^\top \bbeta))$, with
$\bbeta = (-0.4,\, 1.0,\, -0.8,\, 0.6,\, 0.0,\, 0.0)^\top$. The boundary
statistic is $B_{i,\mathrm{diff}} = 1 - 2|\expit(\X_i^\top \bbeta) - 0.5|$.

Three latent difficulty channels are
\begin{align*}
  D_i &= \delta_0 + \delta_1 B_{i,\mathrm{diff}}
        + \delta_2 |X_{i,1}| + \delta_3 |X_{i,2}| + U_i, \quad U_i \sim N(0, \sigma_D^2),\\
  H_i &= \eta_1 X_{i,3} + \eta_2 X_{i,4} + \xi_i, \quad \xi_i \sim N(0, \sigma_H^2),\\
  M_i &= \zeta_1 X_{i,4} + \zeta_2 X_{i,5} + \nu_i, \quad \nu_i \sim N(0, \sigma_M^2).
\end{align*}
The surrogate is
$Q_i = \expit(\tau_0 + \tau_1 \Ystar_i + \tau_2 X_{i,1} + \tau_3 X_{i,2}
- \lambda_D D_i - \lambda_M M_i + \varepsilon_i)$ with
$\varepsilon_i \sim N(0, \sigma_Q^2)$. Coder $j$ has sensitivity intercept
$\alpha^{(1)}_j \sim N(a_{\text{sens}}, \sigma_{\text{coder}}^2)$ and
specificity intercept
$\alpha^{(0)}_j \sim N(a_{\text{spec}}, \sigma_{\text{coder}}^2)$;
conditional on $\Ystar_i$ and the difficulty channels,
$\PR(G_{ij} = 1 \mid \Ystar_i = 1) = \expit(\alpha^{(1)}_j - \gamma_D D_i - \gamma_H H_i)$
and the symmetric expression on $\Ystar_i = 0$. The audit indicator is
uniform Bernoulli($\pi$); audited units draw two distinct coders without
replacement. Under uniform adjudication $\rho_i \equiv \bar\rho$; under
disagreement-based adjudication $\rho_i \propto d_0 + d_1
\one\{G_{i,1} \neq G_{i,2}\}$ rescaled to mean $\bar\rho$.

\begin{table}[h]
  \caption{Scenario parameters varied across the synthetic sweep.
    Default values not listed: $\delta_0 = 0$, $\delta_1 = 1.5$,
    $\delta_2 = \delta_3 = 0.4$, $\sigma_D = 0.75$, $\eta_1 = 0.8$,
    $\eta_2 = -0.5$, $\sigma_H = 0.5$, $\zeta_1 = 0.6$, $\zeta_2 = 0.5$,
    $\sigma_M = 0.5$, $\tau_0 = 0$, $\tau_2 = 0.3$, $\tau_3 = -0.2$,
    $a_{\text{sens}} = 2.4$, $a_{\text{spec}} = 2.6$, $J = 8$,
    $d_0 = 0.1$, $d_1 = 1.0$.}
  \label{tab:appx-scenarios}
  \centering
  \small
  \begin{tabular}{lccc}
    \toprule
    Parameter & Benign & Realistic & Hard \\
    \midrule
    $\tau_1$              & 2.0 & 1.5 & 1.0 \\
    $\lambda_D$           & 0.5 & 1.0 & 1.4 \\
    $\lambda_M$           & 0.4 & 0.8 & 1.2 \\
    $\sigma_Q$            & 0.7 & 1.0 & 1.4 \\
    $\sigma_{\text{coder}}$ & 0.20 & 0.35 & 0.50 \\
    $\gamma_D$            & 0.5 & 1.0 & 1.4 \\
    $\gamma_H$            & 0.35 & 0.7 & 1.0 \\
    Adjudication policy   & uniform & disagreement & disagreement \\
    $\bar\rho$            & 0.25 & 0.25 & 0.10 \\
    \bottomrule
  \end{tabular}
\end{table}

\section{Inner-Stage Feature Ablation}
\label{appx:ablation}

\Cref{tab:appx-ablation} ablates the inner nuisance feature set
($R = 200$ replications). Coder-pair features add no detectable
improvement over the baseline feature set in any scenario. We adopt the
baseline-feature logistic variant as the default \PADSL{} throughout the
main text.

\begin{table}[h]
  \caption{Inner-stage feature ablation, $\beta_1$ ($R = 200$ replications,
    true $\beta_1 = 1$). \textbf{Bold} marks the column-best for MC SD,
    RMSE, and CI length among the design-valid estimators
    (\textsc{DSL-AdjOnly}, \PADSL{} variants); coverage is a calibration
    diagnostic and bias is reported but not bolded.}
  \label{tab:appx-ablation}
  \centering
  \small
  \begin{tabular}{llrrrrr}
    \toprule
    Scenario & Inner stage & Bias & MC SD & RMSE & Coverage & CI len. \\
    \midrule
    benign & \textsc{DSL-AdjOnly} (no inner)    & $0.006$ & $0.057$ & $0.057$ & $0.965$ & $0.213$ \\
           & \PADSL{} (logistic, baseline)      & $0.005$ & $\best{0.047}$ & $\best{0.047}$ & $0.950$ & $\best{0.173}$ \\
           & \PADSL{} (logistic, +coder)        & $0.006$ & $0.048$ & $0.048$ & $0.935$ & $0.174$ \\
    \midrule
    realistic & \textsc{DSL-AdjOnly} (no inner) & $0.027$ & $0.130$ & $0.133$ & $0.960$ & $0.524$ \\
           & \PADSL{} (logistic, baseline)      & $0.015$ & $\best{0.116}$ & $\best{0.117}$ & $0.945$ & $\best{0.466}$ \\
           & \PADSL{} (logistic, +coder)        & $0.016$ & $0.118$ & $0.118$ & $0.940$ & $0.469$ \\
    \midrule
    hard   & \textsc{DSL-AdjOnly} (no inner)    & $0.052$ & $\best{0.248}$ & $\best{0.253}$ & $0.950$ & $\best{0.932}$ \\
           & \PADSL{} (logistic, baseline)      & $0.051$ & $0.250$ & $0.255$ & $0.945$ & $\best{0.932}$ \\
           & \PADSL{} (logistic, +coder)        & $0.052$ & $0.251$ & $0.255$ & $0.950$ & $0.948$ \\
    \bottomrule
  \end{tabular}
\end{table}

\section{Detox: Additional Proxy Truths}
\label{appx:detox-extra}

\Cref{tab:appx-detox-extra} reports the Detox $\beta_1$ comparison under the
two proxy-truth constructions not shown in the main text: hard majority vote
on the reference panel ($R = 50$ replications, target $\beta_1 = -1.50$) and
the soft mean attack share ($R = 50$, target $\beta_1 = -1.10$). The
qualitative ranking is unchanged: the plug-in baselines and \textsc{DSL-MajVote}
fed with noisy gold miss the proxy with zero coverage on the hard-majority
proxy; \textsc{DSL-AdjOnly} and \PADSL{} both achieve
nominal coverage; \PADSL{} reduces RMSE relative to \textsc{DSL-AdjOnly} by
$13$--$16\%$. Because $\Ystar$ is unobservable in the field, the displayed
\emph{truth} is the proxy refit on the full reference panel;
\textsc{DSL-oracle} is therefore an upper bound on achievable accuracy
under each proxy, not an oracle for the latent attack construct.

\begin{table}[h]
  \caption{Detox semi-synthetic study, $\beta_1$ on \texttt{logged\_in}
    under the two proxy-truth constructions deferred from the main text.
    \textbf{Bold} marks the column-best for MC SD, RMSE, and CI length per
    proxy among design-valid estimators (\textsc{DSL-AdjOnly}, \PADSL{});
    coverage is a calibration diagnostic and bias is reported but not
    bolded. The light rule separates the infeasible \textsc{DSL-oracle}
    benchmark.}
  \label{tab:appx-detox-extra}
  \centering
  \small
  \begin{tabular}{llrrrrr}
    \toprule
    Proxy & Method & Bias & MC SD & RMSE & Coverage & CI len. \\
    \midrule
    hard maj. & \textsc{Surrogate-only} & $\phantom{-}0.393$ & $0.000$ & $0.393$ & $0.00$ & $0.050$ \\
              & \textsc{Human-naive}    & $\phantom{-}0.409$ & $0.046$ & $0.412$ & $0.00$ & $0.208$ \\
              & \textsc{DSL-MajVote}    & $\phantom{-}0.416$ & $0.039$ & $0.418$ & $0.00$ & $0.180$ \\
              & \textsc{DSL-AdjOnly}    & $-0.005$           & $0.102$ & $0.101$ & $0.98$ & $0.443$ \\
              & \PADSL{}                & $-0.006$           & $\best{0.088}$ & $\best{0.087}$ & $0.98$ & $\best{0.398}$ \\
    \oraclesep
              & \textsc{DSL-oracle}     & $-0.000$           & $0.047$ & $0.046$ & $0.96$ & $0.179$ \\
    \midrule
    soft mean & \textsc{Surrogate-only} & $\phantom{-}0.285$ & $0.000$ & $0.285$ & $0.00$ & $0.035$ \\
              & \textsc{Human-naive}    & $\phantom{-}0.009$ & $0.046$ & $0.046$ & $0.98$ & $0.208$ \\
              & \textsc{DSL-MajVote}    & $\phantom{-}0.017$ & $0.038$ & $0.041$ & $1.00$ & $0.179$ \\
              & \textsc{DSL-AdjOnly}    & $\phantom{-}0.013$ & $0.067$ & $0.068$ & $0.90$ & $0.259$ \\
              & \PADSL{}                & $\phantom{-}0.012$ & $\best{0.056}$ & $\best{0.057}$ & $0.98$ & $\best{0.225}$ \\
    \oraclesep
              & \textsc{DSL-oracle}     & $\phantom{-}0.003$ & $0.019$ & $0.019$ & $1.00$ & $0.097$ \\
    \bottomrule
  \end{tabular}
\end{table}

\section{Design Curves: Audit and Adjudication Sweep}
\label{appx:design-curves}

The three benign / realistic / hard scenarios in the main text are static
design points. To answer the practical question ``\emph{how much
adjudication should an analyst buy, and how should adjudicated items be
selected?}", we sweep the audit rate $\pi \in \{0.05, 0.10, 0.20\}$,
the mean adjudication rate $\bar\rho \in \{0.10, 0.25, 0.50\}$, the
adjudication policy $\in \{$uniform, disagreement$\}$, and an optional
positivity floor $\rho_{\min,\text{agree}} \in \{0, 0.05\}$ on agreement
cases (under disagreement-driven adjudication, units with
$G_{i,1} = G_{i,2}$ would otherwise receive
$\rho \approx d_0/(d_0 + d_1)\,\bar\rho$, which can be very small).
Other parameters are fixed at the realistic-scenario values
(\cref{tab:appx-scenarios}); each cell uses $R = 200$ replications.

\begin{figure}[h]
  \centering
   \includegraphics[width=\linewidth]{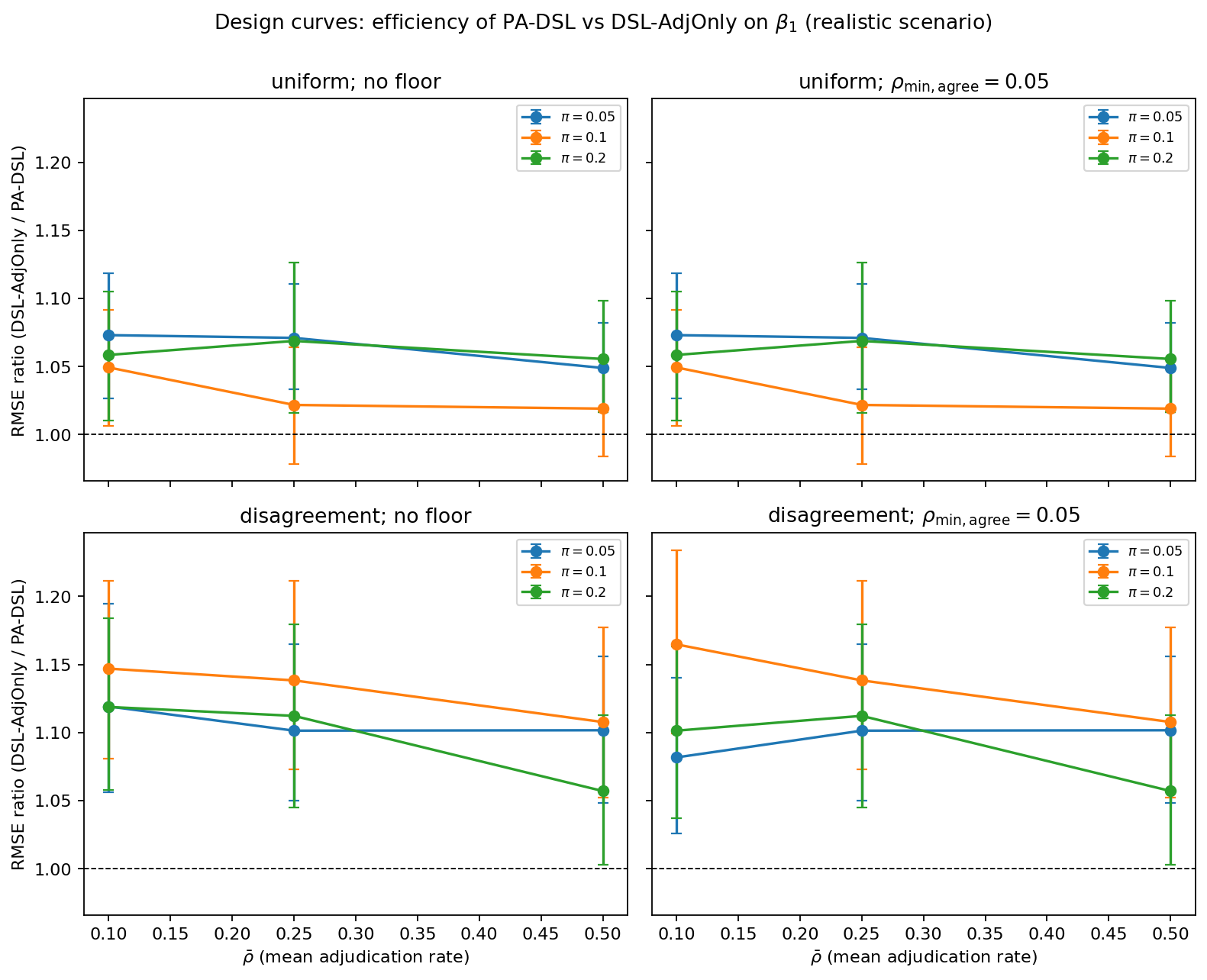}%
  \caption{Design-curve sweep on the realistic scenario. RMSE ratio
    $\mathrm{RMSE}(\textsc{DSL-AdjOnly}) / \mathrm{RMSE}(\PADSL{})$ for
    $\beta_1$, with $95\%$ paired-bootstrap CIs ($R = 200$ reps per cell,
    $1000$ bootstrap draws). Higher values favor \PADSL{}. The dashed line
    at $1.0$ marks the no-gain point.}
  \label{fig:design-curves}
\end{figure}

\Cref{fig:design-curves} reports the $\beta_1$ RMSE ratio
$\mathrm{RMSE}(\textsc{DSL-AdjOnly}) / \mathrm{RMSE}(\PADSL{})$ across the
sweep, with paired-bootstrap $95\%$ CIs. Three patterns emerge.
\emph{First, disagreement-driven adjudication is consistently higher in
point estimate than uniform adjudication} across the grid: the disagreement
panels (bottom row) sit roughly $5$--$10$ percentage points above the
uniform panels at matched mean budget, with overlapping but offset CIs at
$R = 200$ replications per cell. The mechanism is that adjudication
concentrates on the cases where coder labels disagree --- exactly where
the inner $\hat\mu$ has the largest residual to debias and the most
leverage to add over \textsc{DSL-AdjOnly}.
\emph{Second, the gain is largest at the moderate audit rate $\pi = 0.10$
under disagreement} (orange line, peak ratio $1.15$ at $\bar\rho = 0.10$);
$\pi = 0.05$ leaves too little inner training data, while $\pi = 0.20$
already buys enough adjudicated truth that the inner correction has less
to add and the ratio drops toward $1.06$ at $\bar\rho = 0.50$.
\emph{Third, the positivity floor on agreement cases has only small,
mostly within-CI effects} on the realistic-scenario sweep. The
$\rho_{\min,\text{agree}} = 0.05$ panels are nearly indistinguishable from
the no-floor panels. The floor would matter more at scenarios with very
small $d_0/(d_0 + d_1)$ ratios (where the unfloored agreement-case
$\rho_i$ is close to zero); we leave that exploration to future work. For
practical pipelines, the sweep suggests targeting $\pi \in [0.10, 0.20]$
with disagreement-driven adjudication at $\bar\rho \in [0.10, 0.25]$ as a
sweet spot of audit cost vs.\ inferential efficiency.

\section{Inner-Stage AIPW Ablation}
\label{appx:inner-ablation}

\PADSL{}'s inner stage combines an outcome model $\hat\mu(Z)$ with a
Horvitz--Thompson correction $V/\rho$ around it. We isolate the contribution of each piece by re-running the synthetic analysis with
inner-stage variants:
\begin{itemize}[leftmargin=2em, itemsep=2pt, topsep=4pt]
  \item \textbf{IPW-only.} Set $\hat\mu \equiv 0$, so
    $\widehat M_i = (V_i/\rho_i)\,A_i$. Pure Horvitz--Thompson at the inner
    stage. Design-valid (unbiased for any nuisance choice) but does not
    exploit the surrogate or coder labels at the inner stage, and is
    sensitive to small $\rho_i$ values.
  \item \textbf{IPW-only (trunc).} As IPW-only but with the truncated
    weight $\rho_i \mapsto \max(\rho_i, 0.05)$, capping $1/\rho_i \le 20$.
    This is a numerical sensitivity check that disentangles weight-stability
    failures from fundamental design issues; truncation breaks
    design-validity under the original sampling weights.
  \item \textbf{Reg-only.} Skip the $V/\rho$ debias; set
    $\widehat M_i = \hat\mu(Z_i)$. Pure regression. Lower variance than
    IPW-only when $\hat\mu$ fits well, but the sandwich SE no longer
    accounts for the nuisance estimation noise, so confidence intervals
    can severely undercover.
\end{itemize}
\Cref{tab:appx-inner-ablation} reports the comparison on the three
synthetic scenarios at $R = 200$.

The IPW-only variant pays a steep variance price for not exploiting the
inner features. In benign and realistic, RMSE roughly triples; in hard the
estimator becomes \emph{numerically unstable}. The disagreement-weighted
adjudication policy at $\bar\rho = 0.10$ produces audited units with
$\rho_i$ as low as $\sim\!0.02$ ($1/\rho_i \approx 45$), and the
$\widetilde Y_i$ pseudo-outcomes inherit those weights. Two of $R = 200$
hard-scenario replications produce pseudo-outcome distributions extreme
enough to push the logistic IRLS solver outside its convergence regime,
yielding $|\hat\beta_1|$ on the order of $10^5$--$10^6$ and dominating the
column means.\footnote{The truncated variant
\textsc{PA-DSL-IPW-only-trunc} caps $1/\rho_i \le 20$ and restores
numerical stability on the hard scenario at the cost of a small bias
relative to the original design. We report it as a sensitivity check that
disentangles a small-$\rho$ stability artefact from a fundamental design
defect; under the original (un-truncated) sampling weights, IPW-only is
the design-valid pure-Horvitz--Thompson baseline.}

The Reg-only variant achieves lower MC variance than full \PADSL{} in
benign and realistic, and looks nearly unbiased, but the empirical
coverage drops to $0.645$ in benign, $0.260$ in realistic, and $0.070$ in
hard. The sandwich SE on $(\X_i, \widetilde Y_i)$ is computed as if
$\widehat M_i = \hat\mu(Z_i)$ were a fixed quantity, but $\hat\mu$ is
itself a noisy estimator; without the $V/\rho$ correction term to absorb
that noise into the estimating equation, the SE underestimates the true
variability and CIs miss the truth far more often than nominal.

The full \PADSL{} estimator avoids both failure modes: the regression
component $\hat\mu(Z)$ stabilizes the inner pseudo-label relative to
IPW-only (cutting variance by an order of magnitude), while the $V/\rho$
correction restores design validity and makes the sandwich variance
reflect the adjudication-stage uncertainty.

\begin{table}[h]
  \caption{Inner-stage AIPW ablation, $\beta_1$ across the three synthetic
    scenarios ($R = 200$ replications, true $\beta_1 = 1$).
    \emph{IPW-only:} inner $\hat\mu \equiv 0$.
    \emph{IPW-only (trunc):} IPW-only with truncated weights $1/\rho \le 20$
    --- a numerical sensitivity check that breaks design-validity under the
    original sampling weights, included only to disentangle weight-stability
    failures from fundamental design issues.
    $^\dagger$ The truncation engages only when some $\rho_i < 0.05$, which
    happens only on the hard scenario; on benign and realistic the
    IPW-only-trunc row reproduces IPW-only verbatim.
    \emph{Reg-only:} no $V/\rho$ debias at the inner stage. \textbf{Bold}
    marks the column-best for MC SD, RMSE, and CI length per scenario among
    design-valid estimators (excluding Reg-only and IPW-only-trunc, which
    are design-invalid by construction); coverage is a calibration
    diagnostic and bias is reported but not bolded. The light rule
    separates the infeasible \textsc{DSL-oracle} benchmark.}
  \label{tab:appx-inner-ablation}
  \centering
  \small
  \begin{tabular}{llrrrrr}
    \toprule
    Scenario & Inner stage & Bias & MC SD & RMSE & Coverage & CI len. \\
    \midrule
    benign & \textsc{DSL-AdjOnly}        & $\phantom{-}0.006$ & $0.057$ & $0.057$ & $0.965$ & $0.213$ \\
           & \PADSL{} (full)             & $\phantom{-}0.005$ & $\best{0.047}$ & $\best{0.047}$ & $0.950$ & $\best{0.173}$ \\
           & \PADSL{} (IPW-only)         & $\phantom{-}0.020$ & $0.139$ & $0.140$ & $0.970$ & $0.559$ \\
           & \PADSL{} (IPW-only, trunc)$^\dagger$ & $\phantom{-}0.020$ & $0.139$ & $0.140$ & $0.970$ & $0.559$ \\
           & \PADSL{} (Reg-only)         & $\phantom{-}0.001$ & $0.046$ & $0.046$ & $0.645$ & $0.079$ \\
    \oraclesep
           & \textsc{DSL-oracle}         & $\phantom{-}0.006$ & $0.029$ & $0.030$ & $0.935$ & $0.110$ \\
    \midrule
    realistic & \textsc{DSL-AdjOnly}     & $\phantom{-}0.027$ & $0.130$ & $0.133$ & $0.960$ & $0.524$ \\
              & \PADSL{} (full)          & $\phantom{-}0.015$ & $\best{0.116}$ & $\best{0.117}$ & $0.945$ & $\best{0.466}$ \\
              & \PADSL{} (IPW-only)      & $\phantom{-}0.085$ & $0.285$ & $0.296$ & $0.970$ & $1.115$ \\
              & \PADSL{} (IPW-only, trunc)$^\dagger$ & $\phantom{-}0.085$ & $0.285$ & $0.296$ & $0.970$ & $1.115$ \\
              & \PADSL{} (Reg-only)      & $-0.007$ & $0.066$ & $0.066$ & $0.260$ & $0.053$ \\
    \oraclesep
              & \textsc{DSL-oracle}      & $\phantom{-}0.009$ & $0.039$ & $0.040$ & $0.925$ & $0.148$ \\
    \midrule
    hard   & \textsc{DSL-AdjOnly}        & $\phantom{-}0.052$ & $\best{0.248}$ & $\best{0.253}$ & $0.950$ & $\best{0.932}$ \\
           & \PADSL{} (full)             & $\phantom{-}0.051$ & $0.250$ & $0.255$ & $0.945$ & $\best{0.932}$ \\
           & \PADSL{} (IPW-only)         & \multicolumn{5}{c}{\emph{numerically unstable; see footnote in prose}} \\
           & \PADSL{} (IPW-only, trunc)  & $-0.258$ & $0.176$ & $0.312$ & $0.655$ & $0.695$ \\
           & \PADSL{} (Reg-only)         & $-0.014$ & $0.123$ & $0.123$ & $0.070$ & $0.023$ \\
    \oraclesep
           & \textsc{DSL-oracle}         & $\phantom{-}0.010$ & $0.042$ & $0.043$ & $0.945$ & $0.158$ \\
    \bottomrule
  \end{tabular}
\end{table}

\section{Shared-Partition vs.\ \texorpdfstring{$K^2$}{K2}-fold Cross-fit}
\label{appx:k2-sanity}

\Cref{prop:can}(b) covers two implementations: leakage-free nested
cross-fitting and the shared-partition \cref{alg:ngdsl} under stability
condition \cref{eq:sp}. \Cref{tab:appx-k2} reports a paired sanity check on
the realistic scenario ($N=50{,}000$, $\bar\rho=0.25$, disagreement-weighted
adjudication, $R=200$ replications, paired per-rep seeds): the two
implementations give the same point estimate to within Monte Carlo error,
consistent with \cref{eq:sp} holding for the fixed-hyperparameter
L2-regularized logistic nuisances used here.

\begin{table}[h]
  \caption{Shared-partition vs.\ $K^2$-fold double cross-fit, $\beta_1$ on
    the realistic scenario ($R = 200$ paired replications). The
    ``mean abs.\ diff.'' column is the per-replication
    $|\hat\beta_1^{\mathrm{sp}} - \hat\beta_1^{\mathrm{K^2}}|$ averaged
    across replications.}
  \label{tab:appx-k2}
  \centering
  \small
  \begin{tabular}{lrrrrrr}
    \toprule
    Cross-fit scheme & Bias & MC SD & RMSE & Coverage & CI len. & Mean abs.\ diff. \\
    \midrule
    Shared partition (\cref{alg:ngdsl}) & $0.0148$ & $0.1160$ & $0.1166$ & $0.945$ & $0.4664$ & --- \\
    $K^2$-fold double cross-fit         & $0.0148$ & $0.1159$ & $0.1165$ & $0.945$ & $0.4666$ & $0.0009$ \\
    \bottomrule
  \end{tabular}
\end{table}

The mean absolute paired difference $0.0009$ is two orders of magnitude
smaller than either implementation's MC SD ($\approx 0.116$), and the
sample correlation between the two paired estimates is $0.99995$. The
shared-partition shortcut reproduces the leakage-free estimator essentially
exactly at the cost of $K - 1$ fewer inner fits per outer fold.

\section{Imperfect-Adjudication Sensitivity}
\label{appx:imperfect-adj}

\cref{ass:correct} ($A_i = \Ystar_i$ on adjudicated units) is load-bearing
for design-validity with respect to $\Ystar_i$. To illustrate what happens
when adjudication is systematically wrong, \cref{tab:appx-biased-adj}
compares the realistic scenario ($N=50{,}000$, $R=500$ replications) under
correct adjudication against a biased-adjudication mechanism that flips
$A_i$ with probability $0.075$ when $X_{i,1} > 0$ and $0.025$ otherwise ---
a 5\% average error correlated with the regression's primary covariate.

Under correct adjudication, both \textsc{DSL-AdjOnly} and \PADSL{} are
unbiased and have nominal coverage. Under biased adjudication, both inherit
the adjudicator bias relative to $\Ystar_i$ and miss the truth at higher
rates. \PADSL{} retains its design-validity for the \emph{adjudicated
construct} (the function $\E[A_i \mid \X_i]$), but that construct is no
longer $\E[\Ystar_i \mid \X_i]$ when adjudicators systematically disagree
with the latent label. The audit/adjudication design probabilities cannot
recover information adjudicators do not provide; designs that further
relax \cref{ass:correct} require an additional measurement tier (e.g.\
treating $A_i$ as a noisy proxy of $\Ystar_i$ with a known design weight).

\begin{table}[h]
  \caption{Imperfect-adjudication sensitivity, $\beta_1$ on the realistic
    scenario ($R = 500$ replications, target $\beta_1 = 1$).
    \emph{Correct A=Y*}: \cref{ass:correct} holds. \emph{Biased A}:
    $A_i$ flipped with probability $0.075$ when $X_{i,1} > 0$ else
    $0.025$. \textbf{Bold} marks the column-best for MC SD, RMSE, and CI
    length within each adjudication-target block among design-valid
    estimators (\textsc{DSL-AdjOnly}, \PADSL{}). The light rule separates
    the infeasible \textsc{DSL-oracle} benchmark.}
  \label{tab:appx-biased-adj}
  \centering
  \small
  \begin{tabular}{llrrrrr}
    \toprule
    Adjudication target & Method & Bias & MC SD & RMSE & Coverage & CI len. \\
    \midrule
    Correct $A = \Ystar$ & \textsc{DSL-AdjOnly} & $\phantom{-}0.020$ & $0.135$ & $0.136$ & $0.952$ & $0.522$ \\
                         & \PADSL{}             & $\phantom{-}0.013$ & $\best{0.123}$ & $\best{0.123}$ & $0.940$ & $\best{0.467}$ \\
    \oraclesep
                         & \textsc{DSL-oracle}  & $\phantom{-}0.007$ & $0.037$ & $0.038$ & $0.946$ & $0.148$ \\
    \midrule
    Biased $A$           & \textsc{DSL-AdjOnly} & $-0.124$           & $0.131$ & $0.180$ & $0.790$ & $0.508$ \\
                         & \PADSL{}             & $-0.129$           & $\best{0.122}$ & $\best{0.178}$ & $0.784$ & $\best{0.472}$ \\
    \oraclesep
                         & \textsc{DSL-oracle}  & $\phantom{-}0.007$ & $0.037$ & $0.038$ & $0.946$ & $0.148$ \\
    \bottomrule
  \end{tabular}
\end{table}

\section{Classification-oriented predictor diagnostics.}
\label{app:classification-diag}
Although PA-DSL targets downstream estimating equations rather than individual classification, we report supplemental prediction diagnostics to help interpret the efficiency mechanism. Table~\ref{tab:predictor-diagnostics} compares the full-frame surrogate \(Q_i\), the outer nuisance prediction \(\widehat g(B_i)\), and the inner nuisance prediction \(\widehat\mu(Z_i)\), evaluated against the adjudicated/proxy target. The final AIPW pseudo-outcomes are not included as classifiers because they are estimating-equation devices and need not be calibrated probabilities or lie in \([0,1]\).

The diagnostics support the mechanism described in Appendix \ref{appx:theory}. In the benign and realistic synthetic settings, \(\widehat\mu(Z_i)\) improves prediction relative to \(\widehat g(B_i)\), indicating that the audit-tier features add adjudication-relevant information beyond \(B_i=(Q_i,X_i)\). In the hard setting, this incremental signal is small and the inner prediction does not improve on \(\widehat g(B_i)\), matching the near-zero efficiency gain reported in the main results. The Detox study shows a modest but consistent improvement from \(\widehat\mu(Z_i)\), consistent with the observed PA-DSL efficiency gain.

\begin{table}[t]
\centering
\caption{Classification-oriented diagnostics for predictive components. AUC is higher-is-better; Brier score is lower-is-better. The RMS difference is \(\{n^{-1}\sum_i(\widehat\mu(Z_i)-\widehat g(B_i))^2\}^{1/2}\), a descriptive measure of incremental audit-tier signal. Diagnostics are evaluated against the adjudicated/proxy target and are not used as inferential validity criteria.}
\label{tab:predictor-diagnostics}
\begin{tabular}{lccccccc}
\toprule
& \multicolumn{2}{c}{\(Q_i\)} 
& \multicolumn{2}{c}{\(\widehat g(B_i)\)}
& \multicolumn{2}{c}{\(\widehat\mu(Z_i)\)}
&  \\
\cmidrule(lr){2-3}\cmidrule(lr){4-5}\cmidrule(lr){6-7}
Study / scenario 
& AUC & Brier 
& AUC & Brier 
& AUC & Brier 
& RMS\((\widehat\mu-\widehat g)\) \\
\midrule
Synthetic: benign    & 0.949 & 0.118 & 0.968 & 0.070 & 0.990 & 0.037 & 0.300 \\
Synthetic: realistic & 0.776 & 0.206 & 0.864 & 0.150 & 0.869 & 0.146 & 0.156 \\
Synthetic: hard      & 0.639 & 0.292 & 0.810 & 0.176 & 0.794 & 0.182 & 0.064 \\
Detox                & 0.903 & 0.096 & 0.902 & 0.095 & 0.908 & 0.088 & 0.096 \\
\bottomrule
\end{tabular}
\end{table}

\end{document}